\documentclass[10pt,twocolumn,letterpaper]{article}
\PassOptionsToPackage{table}{xcolor}
\usepackage[pagenumbers]{cvpr}

\usepackage{amsmath,amsfonts,bm}

\def\eqref#1{equation~\ref{#1}}

\def\1{\bm{1}}

\DeclareMathAlphabet{\mathsfit}{\encodingdefault}{\sfdefault}{m}{sl}
\SetMathAlphabet{\mathsfit}{bold}{\encodingdefault}{\sfdefault}{bx}{n}

\usepackage{float}
\usepackage{placeins}
\usepackage{multirow}
\usepackage{array}
\newcolumntype{C}[1]{>{\centering\arraybackslash}m{#1}}

\usepackage{microtype}

\usepackage{microtype}
\AtBeginDocument{\raggedbottom
  \typeout{CVPR2027 layout active: microtype; raggedbottom.}}
\usepackage{array,tabularx}
\usepackage{float}
\usepackage{marvosym}
\definecolor{cvprblue}{rgb}{0.21,0.49,0.74}
\usepackage[pagebackref,breaklinks,colorlinks,allcolors=cvprblue]{hyperref}
\def\confName{CVPR}
\def\confYear{2027}

\title{Bench2Drive-Plan: Bridging Privileged Planner and End-to-End Autonomous Driving
under Shared Protocols
}

\author{
    \textbf{Haoran Liu}\textsuperscript{1,2,*}\quad
    \textbf{Shengzhe Wang}\textsuperscript{1,2,*}\quad
    \textbf{Zhiyuan Zhang}\textsuperscript{3}\quad
    \textbf{Ziang Liu}\textsuperscript{1,2}\\
    \textbf{Qifeng Li}\textsuperscript{3}\quad
    \textbf{Shaofeng Zhang}\textsuperscript{4}\quad
    \textbf{Xiaosong Jia}\textsuperscript{1,2,\Letter}\\[3pt]
    \textsuperscript{1}Fudan University\\
    \textsuperscript{2}Shanghai Key Laboratory of Multimodal Embodied AI\\
    \textsuperscript{3}Shanghai Jiao Tong University\\
    \textsuperscript{4}University of Science and Technology of China\\[2pt]
    {\small \textsuperscript{*}Equal contribution.\quad
    \textsuperscript{\Letter}Corresponding author.}
}
\begin{document}

\maketitle

\begin{abstract}
\looseness=-1
Learning-based autonomous driving has developed in two separate communities based on
the input information. Privileged planners receive ground-truth object states and HD maps
while end-to-end models use raw sensor inputs. Despite sharing many similar modeling
ideas, differences in scene inputs, simulation, and metrics make it difficult to compare
results and transfer insights across them. In this work, we introduce
\textbf{Bench2Drive-Plan} to bridge this gap by evaluating privileged planners under
current end-to-end protocols. On Bench2Drive, we implement vectorized, diffusion-based, and
language-based planning baselines. We also implement two raster-based CNN baselines using RGB semantic BEV images and
multi-channel semantic masks, respectively.

Building on this benchmark, we conduct a cross-paradigm study of PLUTO and VAD. By exchanging their input modules, we find that ground-truth scene inputs do not confer a large inherent performance advantage over learned perception. Further experiments examine the role of internal designs in the performance gap and provide additional support for this finding. Besides, We also test whether camera images provide useful information beyond structured scene states. Additional images bring no consistent gains on Bench2Drive-Plan but improve some planners on NAVSIM, suggesting that visual cues may benefit planning in real-world scenes. Together, these studies show how a shared benchmark can benefit the studies of the two communities and we hope Bench2Drive-Plan could pave the way for future studies about synergistically training planner and end-to-end models.Project repository: \url{https://github.com/chtholly3/Bench2Drive-Plan}.
\end{abstract}

\section{Introduction}

Privileged planning and end-to-end driving have largely developed as separate research
communities. The privileged planning community~\citep{cheng2024pluto,zheng2025diffusion}
studies how to learn driving trajectories from structured scene inputs, typically using
ground-truth bounding boxes and HD maps as inputs. The end-to-end driving community
learns driving policies from sensor observations
\citep{chen2024end,hu2023planning,jiang2023vad,yin2026pictura}. Despite addressing similar
problems and sharing modeling ideas~\citep{huang2023gameformer,liao2025diffusiondrive},
their methods are usually developed and evaluated on different datasets and benchmarks.
Although privileged policies have also been studied alongside sensor-based methods in
CARLA~\citep{chen2020learning,zhang2021roach,renz2022plant,
gerstenecker2025plant20exposingbiases}, much of privileged planning research focuses on
nuPlan~\citep{caesar2021nuplan}. End-to-end driving methods are more often evaluated on
Bench2Drive and NAVSIM~\citep{dauner2024navsim,jia2024bench2drive}. These differences in
evaluation settings make comparison and the exchange of designs between the two
communities difficult.

To bridge end-to-end driving and privileged planning, we introduce
\textbf{Bench2Drive-Plan}, a benchmark to evaluate privileged planners under the
same scenarios, simulation and metrics as end-to-end methods. We provide a shared scene
preprocessing pipeline covering the privileged scene information used by mainstream
planners. Each planner adapter then selects
the information needed by its model and converts it into the required input format.
The predicted trajectories are executed by
the official Bench2Drive controller
and scored
using the benchmark metrics~\citep{jia2024bench2drive}.

\begin{figure*}[!t]
  \centering
  \includegraphics[width=\linewidth]{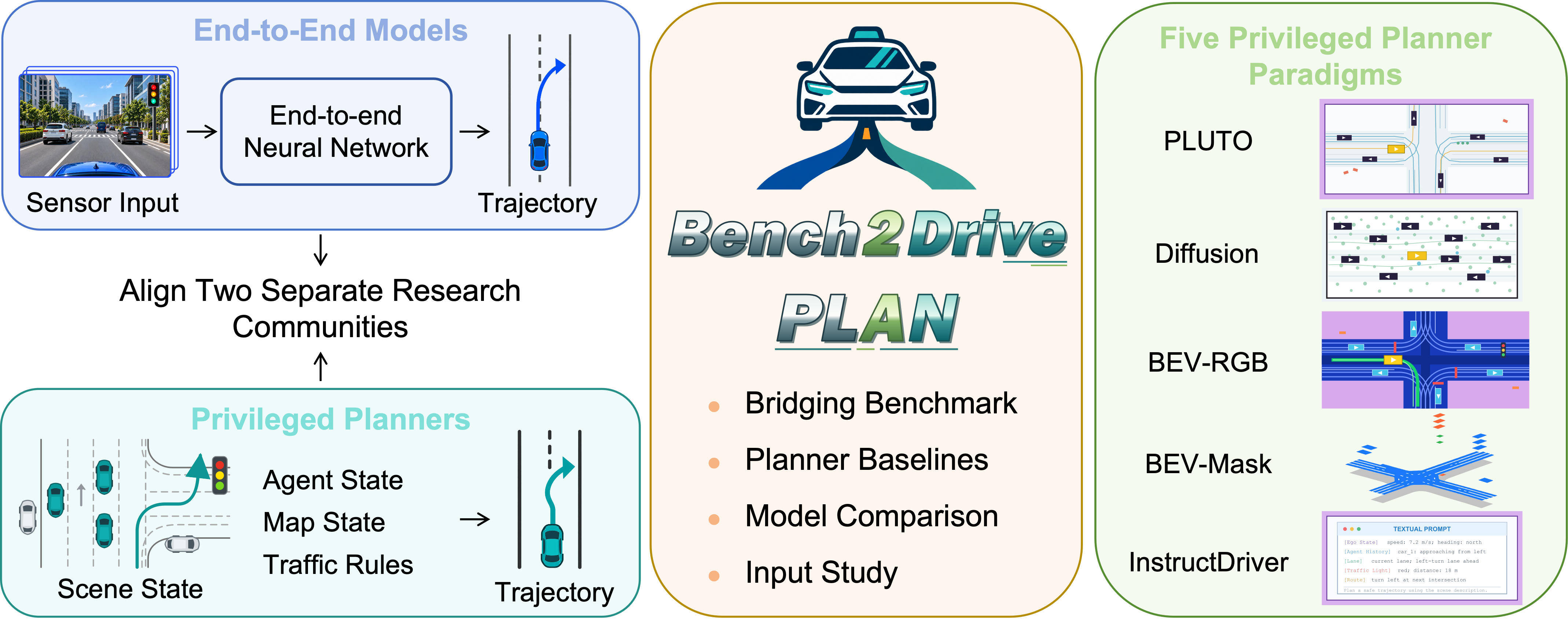}
  \caption{\textbf{Overview of Bench2Drive-Plan.} Our work bridges end-to-end driving and
  privileged planning, evaluates five privileged planners on Bench2Drive-Plan.}
  \label{fig:overview}
\end{figure*}

Bench2Drive-Plan provides five baselines spanning diverse architectures and scene representations for comparison. These include
PLUTO~\citep{cheng2024pluto}, Diffusion Planner~\citep{zheng2025diffusion},
InstructDriver~\citep{zhang2024instruct}, BEV-RGB, and BEV-Mask, covering vectorized,
diffusion-based, language-based, and raster-based planning. The two raster-based CNN
baselines encode scene states as RGB semantic BEV images and multi-channel semantic
masks, respectively~\citep{bansal2018chauffeurnet}.

Our benchmark results show that several sensor-based end-to-end methods achieve
performance comparable to PLUTO despite relying on learned perception rather than
ground-truth scene inputs. To examine whether ground-truth inputs offer a
substantial advantage, we conduct a cross-paradigm experiment that exchanges the input modules
of PLUTO and VAD~\citep{cheng2024pluto,jiang2023vad}. Replacing PLUTO's scene encodings
with VAD perception features reduces its Driving Score (DS) only from 64.74 to 61.37,
while providing VAD with ground-truth scene encodings yields a modest improvement from
44.31 to 48.76. \textbf{These results suggest that ground-truth scene inputs do not provide a large
inherent performance advantage over learned perception in this setting.}
Module connections and internal representations can also affect driving performance
\citep{hu2023planning,weng2024drive,jia2025drivetransformer}. We then
progressively replace PLUTO-style modules with VAD designs to investigate the remaining
performance gap. We identify the performance effects of internal module designs and
improve VAD's DS from 44.31 to 56.38 through changes to modeling and optimization
configurations while retaining its overall architecture, further demonstrating
that the performance gap should not be attributed solely to differences in input
paradigm. This study shows how Bench2Drive-Plan can support comparisons and the exchange
of designs between privileged and end-to-end methods that share similar modeling ideas.

To explore whether visual cues beyond structured scene states help explain the
competitive performance of learned perception, we add additional images input to privileged
planners. We find no consistent
gains on Bench2Drive-Plan, while three of the five planners improve on
NAVSIM~\citep{dauner2024navsim}. These results suggest that additional visual information
may benefit planning in real-world scenes.

We summarize our contributions as follows:
\begin{itemize}[leftmargin=10pt, topsep=0pt, itemsep=1pt, partopsep=1pt, parsep=1pt]
    \item \textbf{Bridging benchmark:} We introduce Bench2Drive-Plan to connect privileged
    planning and end-to-end driving through a shared evaluation protocol. We provide a
    shared scene preprocessing pipeline that extracts and organizes ground-truth scene
    information, allowing planners to access it through model-specific adapters. This
    enables methods previously studied on separate benchmarks to be evaluated under the
    same protocol.

    \item \textbf{Planner baselines:} We adapt PLUTO, Diffusion Planner, and
    InstructDriver, covering established approaches in the planner community. We
    additionally construct BEV-RGB and BEV-Mask, two raster-based CNN baselines that share
    a backbone and training objective but encode scene information differently. These
    implementations broaden the methods evaluated on Bench2Drive and support the study of
    different architectures and scene representations.

    \item \textbf{Cross-paradigm study:} We construct intermediate models between PLUTO
    and VAD to examine the effects of scene inputs and internal designs on planning
    performance. Our experiments show that ground-truth scene inputs do not confer a large
    inherent advantage over learned perception, and that the performance gap should not be
    attributed solely to input paradigms. This study illustrates how Bench2Drive-Plan as a shared benchmark supports comparisons and the exchange of designs between privileged and end-to-end methods.

    \item \textbf{Camera information study:} We investigate whether privileged planners can benefit
    from additional camera images beyond ground-truth scene states. We do not observe the expected improvement on Bench2Drive-Plan, but find an overall improvement on NAVSIM. This study provides a diagnostic perspective on
    the role of visual information in privileged planning.

\end{itemize}

\section{Related Work}

\subsection{End-to-End Driving and Privileged Planning}

End-to-end autonomous driving maps onboard sensor observations to future trajectories or
control
commands \citep{chen2024end,chib2023recent}.
Early studies focused on image-based control and conditional imitation learning
\citep{bojarski2016end,codevilla2018end,kendall2019learning}.
Subsequent methods introduced privileged distillation, intermediate representations, and
multimodal
fusion to improve learning and closed-loop driving
\citep{chen2020learning,prakash2021multimodal,renz2022plant,chen2022learning}.
Recent systems commonly transform camera or LiDAR features into object-centric, BEV, or
occupancy
representations and jointly perform perception, prediction, and planning
\citep{shao2023safety,chitta2022transfuser,li2022bevformer,wu2022trajectory,
hu2023planning,jiang2023vad,
sun2025sparsedrive,zhu2025sparsead,tong2023scene}.

Focusing on the planning stage, existing methods employ diverse scene representations and
trajectory generation strategies.
Rasterized methods encode maps, actors, traffic signals, and routes as semantic BEV inputs
\citep{bansal2018chauffeurnet,phan2020covernet,casas2021mp3}, whereas vectorized methods
represent lane geometry and actor histories as tokens processed by transformers or graph
networks
\citep{gao2020vectornet,huang2023gameformer,zhang2025invdriver}.
PLUTO further combines vectorized scene features with candidate reference lines and robust
imitation learning \citep{cheng2024pluto}.
Diffusion-based planners generate trajectories through conditional denoising and have been
applied to
decision making, control, and autonomous driving planning
\citep{winter2025generative,zhu2023diffusion,zou2025diffusiondrivev2,janner2022planning,
chi2025diffusion,zheng2025diffusion,liao2025diffusiondrive}.
Large language models and vision-language models further formulate driving as a reasoning
problem \citep{li2024large, zhou2024vision, jiang2025survey,
tian2024drivevlmconvergenceautonomousdriving, jiang2026senna}.
Representative methods study coordinate-based trajectory generation, instruction tuning,
language-conditioned driving, visual reasoning, and closed-loop control
\citep{mao2023gpt,zhang2024instruct,shao2024lmdrive,
xu2024drivegpt4,
sima2024drivelm,jia2026bench2drivevl}.

\subsection{Driving Datasets and Simulation Benchmarks}

Autonomous driving benchmarks have evolved from perception datasets such as KITTI
\citep{geiger2012we} to large scale datasets that support perception and prediction,
including nuScenes \citep{caesar2020nuscenes}, Argoverse
\citep{chang2019argoverse,wilson2023argoverse2generationdatasets}, and the Waymo Open
datasets \citep{sun2020scalability,ettinger2021large}.
However, evaluation on recorded data alone does not fully characterize driving behavior
during
execution \citep{chen2024end,li2026ranging}. Existing planning benchmarks provide
different scene interfaces and evaluation protocols.
nuPlan supplies structured scene states and simulation tools commonly used to evaluate
privileged
planners, while NAVSIM evaluates predicted trajectories over logged driving scenes using
the PDM
Score \citep{caesar2021nuplan,dauner2024navsim}.
CARLA provides interactive closed-loop simulation with configurable sensor observations
\citep{dosovitskiy2017carla}.
Built on CARLA, Bench2Drive provides official training data and 44 scenario types covering
diverse
driving capabilities, with recent extensions addressing robustness and speed-conditioned
evaluation
\citep{jia2024bench2drive,zhang2026bench2drive,shao2026can}. 
Bench2Drive-Plan bridges these evaluation ecosystems by benchmarking privileged planners
and
sensor-based end-to-end methods under the same Bench2Drive protocol.

\section{Bench2Drive-Plan}
\begin{figure*}[!t]
    \centering
    \includegraphics[width=\linewidth]{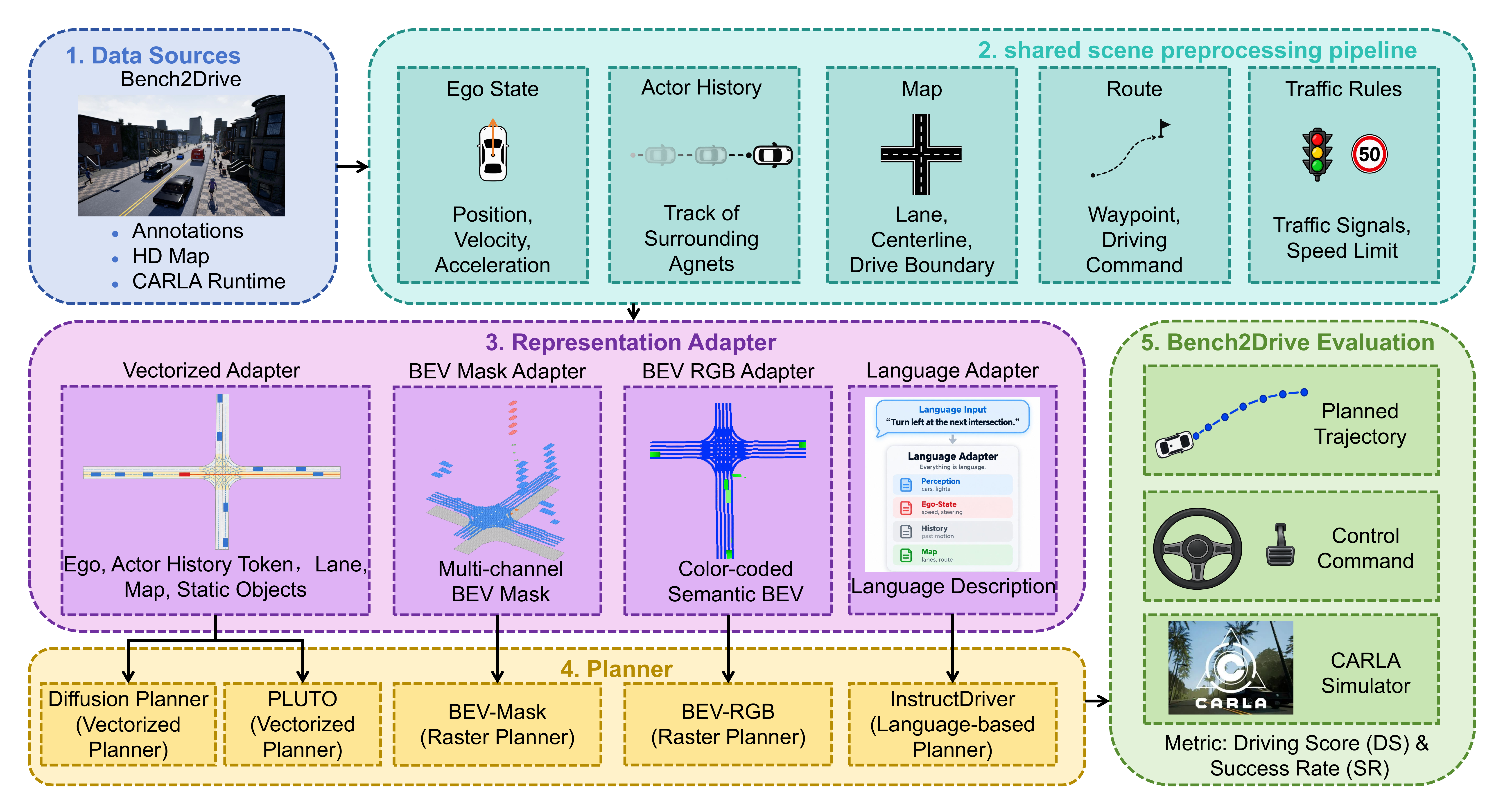}
    \caption{\textbf{Pipeline of the Bench2Drive-Plan.}
  Ground-truth scene states are converted into each planner's input through the shared scene
  preprocessing pipeline and representation-specific adapters, and the predicted ego trajectory is
  executed
  and scored under the official Bench2Drive protocol.}
    \label{fig:framework}
\end{figure*}
\subsection{Framework Overview}

We provide a shared scene preprocessing pipeline that extracts ground-truth scene information
from Bench2Drive, including ego and agent states and histories, static objects, lanes,
crosswalks, traffic-light states, route information, and lane speed limits, covering the
information required by all mainstream privileged planners.

For each frame sample $i$, the pipeline organizes the extracted information into a scene
record $\mathbf{s}_i$. Each privileged planner $m$ implements an adapter $\mathcal{A}_m$
that selects the required fields from $\mathbf{s}_i$ and converts them into its native
input format. The planner $\pi_m$ then predicts a future ego trajectory:
\begin{equation}
    \hat{\tau}_i^{(m)} = \pi_m\!\left(\mathcal{A}_m(\mathbf{s}_i)\right).
\end{equation}

Under this unified framework, we train and evaluate five baseline planners on Bench2Drive-Plan: PLUTO,
Diffusion Planner, BEV-RGB, BEV-Mask and InstructDriver. Together, they span vectorized,
diffusion-based, raster-based and language-based paradigms, enabling systematic comparison
across heterogeneous scene representations and \mbox{model architectures.}

\subsection{Scene Preprocessing}
\label{sec:scene-construction}

The scene preprocessing pipeline extracts scene information for each frame and organizes
it into a shared record. Some fields are read directly, while others require temporal
association, geometric processing, or spatial queries. We use Bench2Drive annotations
during training and synchronized simulator states during closed-loop evaluation. Below, we
describe the construction of actor histories, map features, route information, and traffic
rules.

\textbf{Ego, actors, and static elements.}
For each frame, we extract information about the ego vehicle, other traffic participants,
and static objects from scene annotations or synchronized CARLA states, 
including position, heading, velocity, acceleration, size, and category. We associate observations of the same actor across
frames using persistent actor IDs to construct its history, and transform the geometry and
motion of both current and historical observations to the current ego coordinate frame.
Missing history entries are padded and marked with validity masks. We also preserve the
fine-grained categories needed for Bench2Drive scenarios, including emergency vehicles and
vehicles with an \mbox{open door.}

\textbf{Map preprocessing.}
We reorganize Bench2Drive's nested map data into compact lane and crosswalk records that
can be accessed directly by their identifiers. Each lane stores its centerline, boundaries, and successor connections; crosswalks are represented as polygons. The
processed map is cached for each town. Besides, we build STRtree spatial indices for both lanes and crosswalks, enabling efficient intersection,
containment, and local-range queries~\citep{leutenegger1997str}. 

Most of the planners in our study were originally developed on nuPlan. However,
Bench2Drive contains a substantially larger proportion of long lanes: $22.3\%$ of
Bench2Drive lanes exceed 100\,m, compared with $5.0\%$ in nuPlan. Many planners sample the
same number of points from each lane regardless of its length. Sampling an entire long
lane therefore provides sparse coverage of the geometry near the ego vehicle. When local
lane features are needed, we usually crop the lane polylines around the ego vehicle before
sampling. For reference, Diffusion Planner achieves a Bench2Drive Driving Score of 37.80
using the original lane polylines and 58.78 using locally cropped lanes. Further details
about our lane preprocessing are provided in
Appendix~\ref{app:bench2drive-implementation}.

\textbf{Route and traffic rules.}
Route information specifies the sequence of lanes traversed by the ego vehicle. During
training, we obtain the route directly from the annotations, whereas during evaluation, we
construct a feasible lane-level path from the start to the goal using graph-based search
\citep{tarjan1972depth}. We further augment each lane with traffic-rule information.
Specifically, the traffic-light state is determined from the corresponding trigger region.
Since speed limits are available only at sparse map locations, we index these samples
with a cKDTree, query those near each lane centerline, and assign their median value to
the lane \citep{virtanen2020scipy}.

\subsection{Planner Adapters}\label{sec:adapters}
To adapt the different input format of privileged planners, we construct corresponding
planner adapters.

\textbf{Vectorized planners.} The vectorized adapter converts the extracted scene
information into structured vectors for Diffusion Planner and PLUTO.
 Since the shared scene records already contain numerical state and geometry information, the vectorized adapter mainly selects and organizes these fields into the input formats required by Diffusion Planner~\citep{zheng2025diffusion}
and PLUTO~\citep{cheng2024pluto}. This includes matching their actor capacities, history sampling, and feature definitions. For PLUTO, the adapter additionally constructs candidate reference lines from the route-lane sequence.

\textbf{Raster planners.} Unlike camera-based BEV construction that lifts multi-view
images into a BEV \citep{philion2020lift,huang2021bevdet,li2023bevdepth}, our raster
baselines directly rasterize the ground-truth scene state into a semantic BEV. The BEV-RGB
adapter renders the extracted scene information as a semantic RGB image.
Colors distinguish scene elements, while brightness represents actor history. The BEV-Mask
adapter
instead separates static map structure, actor categories, traffic rules, and sampled
history into
38 binary channels.

\textbf{Language-based planners.} The language adapter converts the scene state into an
ego-centric textual prompt \citep{zhang2024instruct}, providing a structured description
of the relevant scene and traffic context.
The LLM is prompted to generate a reasoning trace and ego-relative future
poses represented as $(x,y,\mathrm{heading})$.
The parser converts these poses to the Bench2Drive trajectory interface.

\subsection{Planner Evaluation Protocol}\label{sec:scoring}

Each method is deployed as a CARLA Leaderboard agent under the same route partitions and
simulator
settings.
At each simulation tick, current and historical CARLA states, route and map records, and
traffic-rule states pass through the scene construction pipeline and the planner adapter
described above to form the model input.
All planners produce trajectories that are executed by the official Bench2Drive
trajectory-tracking controller, which produces throttle, brake, and steering commands. The resulting driving behavior is evaluated using the official Bench2Drive scoring protocol to obtain the final scores.

\section{Benchmark Results}\label{sec:benchmark-results}

\subsection{Benchmark Baselines}
We select the following five privileged planners as baselines:

\textbf{PLUTO}~\citep{cheng2024pluto} encodes agent histories, static obstacles, and
vectorized map elements into a scene representation, then combines lane-derived lateral
queries with learnable longitudinal queries to decode and score multimodal ego-trajectory
proposals.

\textbf{Diffusion Planner}~\citep{zheng2025diffusion} generates trajectories through an
iterative denoising process conditioned on a
vectorized representation of the scene information.

\textbf{BEV-RGB and BEV-Mask}  are constructed as two raster planners. Both use
a ResNet-50~\citep{he2016deep} pretrained on ImageNet~\citep{deng2009imagenet}, followed
by global average pooling and a linear
trajectory head that directly regresses future ego poses. BEV-RGB represents the scene as
a RGB image, using different colors to distinguish scene elements. BEV-Mask represents the
scene information in 38 binary channels that separate semantic categories and sampled
history steps. They are trained with the same imitation
objective.

\textbf{InstructDriver}~\citep{zhang2024instruct} represents structured scene information
as text and uses a language model trained on driving instructions to predict future ego
waypoints.

\textbf{End-to-end Baselines.}
We use VAD~\citep{jiang2023vad},
UniAD-Base~\citep{hu2023planning},
TCP-traj~\citep{wu2022trajectory}, BridgeAD~\citep{zhang2025bridging},
DriveTransformer~\citep{jia2025drivetransformer}, ThinkTwice~\citep{jia2023think},
DriveDPO~\citep{shang2025drivedpo}, WoTE~\citep{li2025end},
DriveAdapter~\citep{jia2023driveadapter},
Drive-JEPA~\citep{wang2026drive}, SimLingo~\citep{renz2025simlingo} and
HiP-AD~\citep{tang2025hip} as twelve
sensor-based end-to-end \mbox{reference baselines.}

\subsection{Experimental Setup}\label{sec:exp}

\paragraph{Bench2Drive evaluation.}
We train the privileged planners on the official Bench2Drive training split, which
contains 950 scenes across 12 towns, and evaluate all five planners on the held-out
220-route base
split. Each method is executed through the official Bench2Drive control and scoring
pipeline. We report Driving Score (DS), which combines route completion
with penalties for driving infractions, and Success Rate (SR), the route completion rate
without a disqualifying infraction. Following the Bench2Drive multi-ability protocol
\citep{jia2024bench2drive}, we additionally report Overtaking, Merging, Emergency Brake,
Give Way,
and Traffic Signs scores and their unweighted mean.
\subsection{Main Results on Bench2Drive}
Table~\ref{tab:b2d-main} reports DS and SR for the five privileged planners and twelve
sensor-based
end-to-end baselines on Bench2Drive. Among the privileged planners, the DS ranking is
PLUTO, Diffusion Planner, InstructDriver, BEV-RGB, and BEV-Mask; SR exhibits
the same broad trend. Among the two matched raster baselines, BEV-RGB performs
substantially better than BEV-Mask, showing how the scene is encoded has a substantial
effect.
\begin{table*}[!t]

\caption{Bench2Drive results and model characteristics. The first five methods are
privileged planners; the remaining twelve are sensor-based end-to-end baselines. Paradigm
indicates the dominant modeling paradigm rather than specific network components. All
methods are evaluated on the official Bench2Drive evaluation routes; $^\ast$ denotes
expert-feature distillation~\citep{jia2024bench2drive}. Venue denotes the original
publication; Ours marks the baselines introduced here.}

\label{tab:b2d-main}

\centering
\footnotesize

\setlength{\tabcolsep}{2.5pt}
\renewcommand{\arraystretch}{1.12}

\begin{tabularx}{\textwidth}{@{}>{\centering\arraybackslash}Xc
*{3}{>{\centering\arraybackslash}X}
*{2}{>{\centering\arraybackslash}m{1.4cm}}@{}}

\toprule
Method & Venue & Input & Paradigm & Output & DS$\uparrow$ & SR$\uparrow$ \\
\midrule

\rowcolor{gray!10}
PLUTO
& arXiv'24
& GT scene vectors
& Transformer
& Trajectory
& \textbf{64.74} & \textbf{35.91} \\

Diffusion Planner
& ICLR'25
& GT scene vectors
& DiT
& Trajectory
& 58.78 & 26.36 \\

\rowcolor{gray!10}
BEV-RGB
& Ours
& GT semantic RGB BEV
& CNN
& Trajectory
& 51.10 & 21.36 \\

BEV-Mask
& Ours
& GT multi-channel BEV
& CNN
& Trajectory
& 41.30 & 13.64 \\

\rowcolor{gray!10}
InstructDriver
& arXiv'24
& GT text prompt
& LLM
& Text trajectory
& 57.16 & 24.09 \\

\midrule

VAD
& ICCV'23
& Multi-view cameras
& Transformer
& Trajectory
& 38.65 & 17.27 \\

\rowcolor{gray!10}
UniAD-Base
& CVPR'23
& Multi-view cameras
& Transformer
& Trajectory
& 38.69 & 19.09 \\

TCP-traj$^\ast$
& NeurIPS'22
& Front camera + ego state
& CNN/MLP/GRU
& Trajectory
& 51.44 & 26.36 \\

\rowcolor{gray!10}
ThinkTwice$^\ast$
& CVPR'23
& Multi-view cameras
& Transformer
& Trajectory
& 52.84 & 27.73 \\

BridgeAD
& CVPR'25
& Multi-view cameras
& Transformer
& Trajectory
& 50.06 & 22.73 \\

\rowcolor{gray!10}
DriveTransformer
& ICLR'25
& Multi-view cameras
& Transformer
& Trajectory
& 60.20 & 35.91 \\

DriveDPO
& NeurIPS'25
& Cameras + LiDAR
& Transformer
& Trajectory
& 62.02 & 30.62 \\

\rowcolor{gray!10}
WoTE
& ICCV'25
& Cameras + LiDAR
& Transformer
& Trajectory
& 61.71 & 31.36 \\

DriveAdapter$^\ast$
& ICCV'23
& Cameras + LiDAR
& Transformer
& Control
& 54.96 & 30.45 \\

\rowcolor{gray!10}
Drive-JEPA
& arXiv'26
& Front camera + ego state
& Transformer
& Trajectory
& 64.52 & 36.82 \\

SimLingo
& CVPR'25
& Multi-view cameras
& VLA
& Trajectory
& 85.07 & 67.27 \\

\rowcolor{gray!10}
HiP-AD
& ICCV'25
& Multi-view cameras
& Transformer
& Trajectory
& \textbf{86.77} & \textbf{69.09} \\

\bottomrule

\end{tabularx}

\end{table*}
PLUTO achieves the highest DS of $64.74$ and SR of $35.91$ among the tested privileged
planners. Several sensor-based end-to-end methods achieve comparable performance despite
relying on learned perception rather than ground-truth scene information. This raises a
question: do planners perform substantially better with accurate ground-truth scene information than with outputs from a learned perception module?
We therefore conduct more controlled experiments in Sec.~\ref{sec:cross-paradigm} to
investigate \mbox{this question.}

\section{Cross-Paradigm Study}
\label{sec:cross-paradigm}

To answer this question, we study PLUTO and VAD, which use different input paradigms but share similar planning
designs. Both represent scene elements as features and generate planning trajectories
through feature interactions, allowing some design choices to be swapped between the two
models.

\subsection{Ground Truth vs. Perception}

\begin{figure*}[!t]
  \centering
  \includegraphics[width=\linewidth]{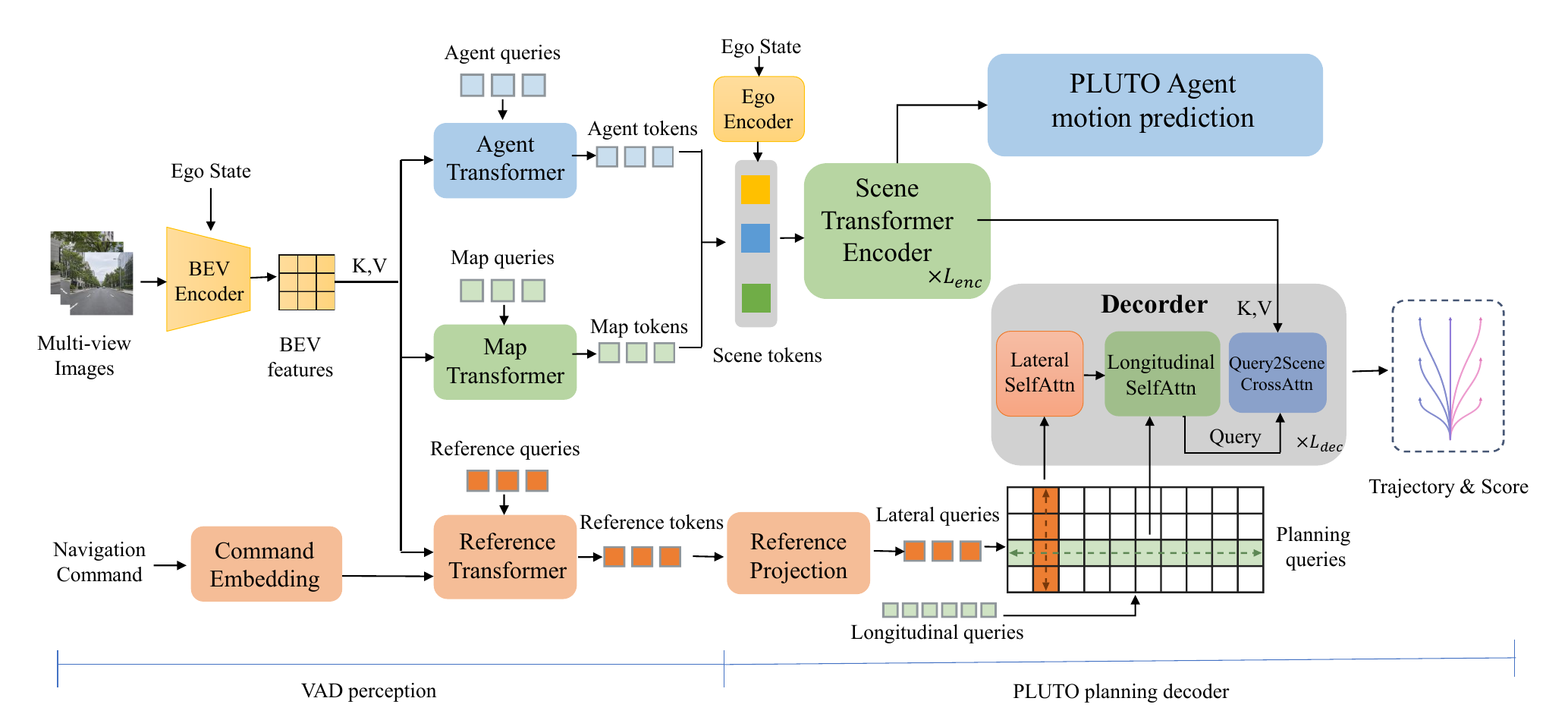}
  \caption{\textbf{PLUTO--VAD architecture.} VAD perception features and an ego-state
  token form the scene context for PLUTO's planning decoder. Navigation-conditioned
  reference queries provide lateral conditioning, combined with learned longitudinal
  mode embeddings. Agent motion prediction is a parallel branch. Static-object tokens
  and feature projection and masking are omitted for clarity.}
  \label{fig:pluto-vad-architecture}
\end{figure*}

We first construct \textbf{PLUTO--VAD}, which replaces PLUTO's structured scene encodings
with camera-derived VAD queries (Figure~\ref{fig:pluto-vad-architecture}). The image
backbone and temporal BEV network produce
agent, map and reference-line queries; these features supply
PLUTO-style scene interaction and reference-conditioned multimodal planning. This model
connects a planning design developed for privileged inputs with a learned perception
representation. Implementation details are provided in
Appendix~\ref{app:cross-paradigm-details}.

As shown in Table~\ref{tab:gt-vs-perception}, PLUTO--VAD and privileged PLUTO achieve similar Driving Scores (DS) of 61.37 and 64.74, respectively. We do not observe the
large performance advantage from ground-truth inputs in this comparison. To further verify this finding, We test the reverse direction by constructing \textbf{VAD--GT}, which retains
VAD's planning components and replaces learned perception features with scene encodings of
ground-truth scene information from our shared scene preprocessing pipeline. VAD--GT achieves
a DS of 48.76, compared with 44.31 for our re-evaluated sensor-based VAD baseline. The improvement is modest, and planning performance remains poor despite access to ground-truth scene information.

\begin{table}[!htbp]
\centering
\small
\setlength{\tabcolsep}{4pt}
\caption{Driving scores with ground-truth scene inputs and learned perception features.
VAD denotes the baseline re-evaluated with the perception module used in this study.}
\label{tab:gt-vs-perception}
\begin{tabular*}{\columnwidth}{@{\extracolsep{\fill}}llc@{}}
\toprule
Model & Scene representation & DS $\uparrow$ \\
\midrule
PLUTO & Ground-truth states & 64.74 \\
PLUTO--VAD & Learned perception & 61.37 \\
\midrule
VAD--GT & Ground-truth states & 48.76 \\
VAD & Learned perception & 44.31 \\
\bottomrule
\end{tabular*}
\end{table}

Together, these results suggest that, In current setting, \textbf{ground-truth inputs do not yield a large performance advantage, while learned perception features still support competitive closed-loop planning.} This suggests that performance differences should not be attributed to perception errors alone and motivates a closer examination of how downstream planning design affects driving performance.

\subsection{From PLUTO--VAD to VAD}

PLUTO--VAD and VAD use nearly identical perception modules, yet achieve DS of 61.37 and
44.31, respectively. This substantial gap motivates us to construct intermediate models
to investigate the design choices behind it.

\begin{figure}[!htbp]
  \centering
  \includegraphics[width=\columnwidth]{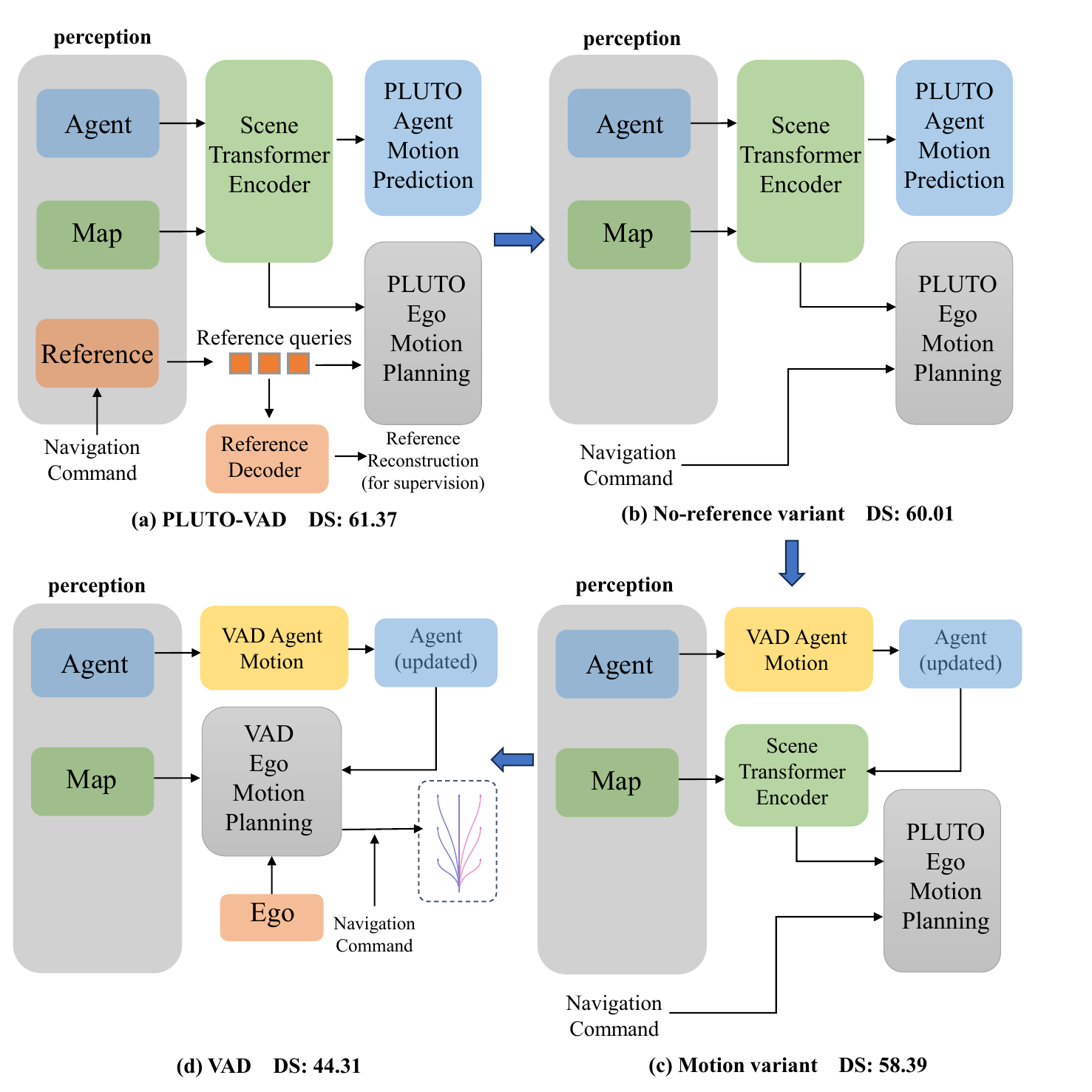}
  \caption{\textbf{Intermediate models between PLUTO--VAD and VAD.}
  The panels show (a) PLUTO--VAD, (b) No-reference, (c) Motion, and (d) VAD.
  DS denotes the Driving Score on Bench2Drive-Plan.}
  \label{fig:pluto-vad-transitions}
\end{figure}

We first consider reference conditioning as a possible explanation for the performance
gap. PLUTO--VAD includes a reference line branch that produces features to initialize its
planning queries. This branch also reconstructs explicit reference lines for auxiliary
supervision. VAD has no corresponding branch. We therefore construct the
\textbf{No-reference variant}, shown in Figure~\ref{fig:pluto-vad-transitions}. It replaces
the reference branch with navigation-command and mode embeddings and removes
lateral self-attention. Its DS changes from 61.37 to 60.01. We next examine how neighboring-agent motion is used in planning. PLUTO
predicts neighboring-agent trajectories in parallel with ego planning, while VAD passes
motion features to its planner. Starting from No-reference, the \textbf{Motion variant}
introduces VAD's agent-agent and agent-map interactions and feeds the updated agent
features into PLUTO's scene encoder and planning decoder. It achieves a DS of 58.39.
After these changes to the major modules, a substantial gap of 14.08 DS remains between
Motion and VAD. This motivates us to investigate whether differences in ego planning
are the main source of the remaining gap. We examine these differences through
ablations on VAD in the next subsection. Further implementation details are provided in
Appendix~\ref{app:cross-paradigm-details}.

\subsection{Modeling and Optimization Choices}

Apart from differences in module design, the planning decoders of PLUTO and VAD
follow similar approaches to feature interaction and trajectory generation, but differ in
some modeling and optimization choices. To investigate whether these choices contribute to
the remaining gap, we conduct ablations on VAD while retaining \mbox{overall architecture.}

We compare two trajectory representations: predicting and supervising per-step
displacements, or directly predicting and supervising future positions in the ego frame.
In the VAD configuration, the planning regression loss has a lower magnitude
than the perception loss. We therefore adjust its normalization to average only over valid
coordinates that receive supervision, increasing the effective scale of the planning loss
in training. PLUTO masks out invalid scene features during attention. Following this
practice, we introduce confidence filtering in VAD's ego attention, retaining only
agent/map queries with classification confidence of at least 0.5 to reduce the influence
of background features on planning interactions.

The loss definitions, confidence calculation, and padding treatment are provided in
Appendix~\ref{app:cross-paradigm-details}.

Table~\ref{tab:endpoint-design} reports the ablation results. Direct position prediction
raises DS from 44.31 to 47.48. On this position-predicting baseline, adjusting the
planning-loss scale and filtering queries individually improve DS to 51.42 and 50.53.
Combining them reaches 56.38, a 12.07-point improvement over VAD and within only two
points of Motion. These changes recover a substantial part of the gap without replacing
VAD's \mbox{overall architecture.}.

\begin{table}[t]
\centering
\small
\setlength{\tabcolsep}{3pt}
\caption{Modeling and optimization choices within VAD. Adjusted loss scale uses
valid-coordinate normalization (Appendix~\ref{app:cross-paradigm-details}). Filtering uses
a confidence threshold of 0.5.}
\label{tab:endpoint-design}
\begin{tabular*}{\columnwidth}{@{\extracolsep{\fill}}lccr@{}}
\toprule
Trajectory output & Loss scale & Filtering & DS $\uparrow$ \\
\midrule
Displacements & Original & No & 44.31 \\
Absolute XY & Original & No & 47.48 \\
Absolute XY & Adjusted & No & 51.42 \\
Absolute XY & Original & Yes & 50.53 \\
Absolute XY & Adjusted & Yes & \textbf{56.38} \\
\bottomrule
\end{tabular*}
\end{table}

\begin{figure*}[!t]
    \centering
    \includegraphics[width=\linewidth]{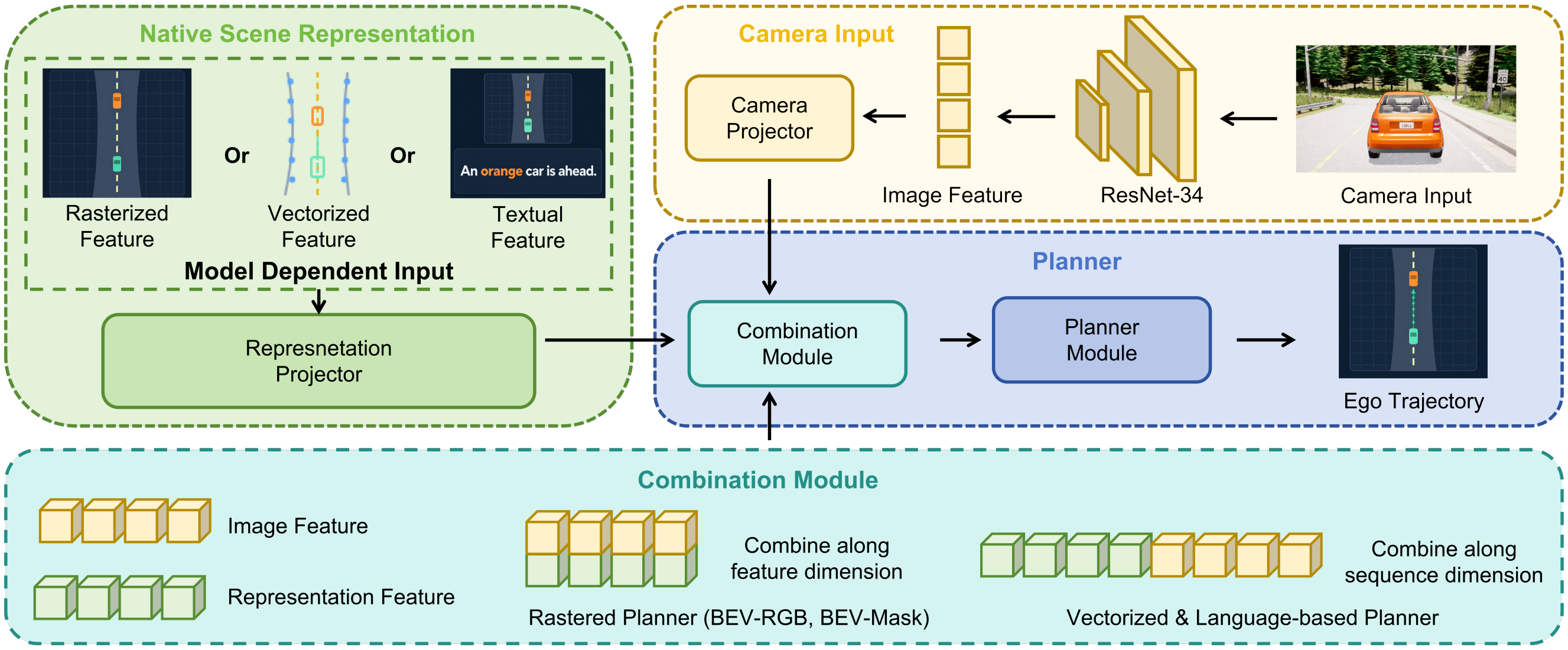}
    \caption{\textbf{Camera fusion.}
  A front-view camera feature is combined with each planner's original scene
  representation.}
    \label{fig:camera_information_study}
\end{figure*}

\subsection{Implications}
The PLUTO--VAD comparison shows that ground-truth scene inputs do not provide a large
inherent advantage over learned perception in this setting. The subsequent ablations help explain the performance differences within this set of models: reference
conditioning and the prediction module account for relatively modest changes along the
bridge, while adjustments within VAD's planning decoder recover a substantial part of its gap to Motion.

This study illustrates one use of Bench2Drive-Plan. A
common evaluation setting allows components to be combined or replaced to investigate how
different design choices affect performance. Our experiments provide a concrete example of
such a comparison across input paradigms.

\section{Camera Information Study}
\label{sec:camera-study}

Our previous study shows that ground-truth scene inputs do not provide an expected large performance
advantage over learned perception. One possible
explanation is that raw camera images contain visual cues beyond explicit scene states,
which may benefit the planners. Bench2Drive-Plan provides both a ground-truth scene
interface and camera images. It therefore offers a platform for studying whether
additional image inputs can improve \mbox{planning performance.}

\subsection{Bench2Drive-Plan Study}

Camera input is integrated into
each planner in its own representation. A ResNet-34 encodes the front-view image
$I_{\mathrm{img}}$, and a learned projection maps its pooled features into the planner's
feature space to obtain the camera representation $\mathbf{c}_m$. Due to differences in
model architectures, planners then consume $\mathbf{c}_m$ differently. PLUTO and Diffusion
Planner append it as an extra token to their scene-token sequence before the fusion
Transformer; BEV-RGB and BEV-Mask concatenate it with the pooled BEV feature and pass the
result through a fusion MLP; InstructDriver prepends it to the text token sequence and
feeds them into the language model. Full details are given in Appendix~\ref{app:camera}.

On Bench2Drive-Plan, additional camera inputs do not produce the expected gains: most planners
exhibit slight degradation (Table~\ref{tab:camera}).

\subsection{Extension to NAVSIM}

Since camera augmentation provides no overall improvement on Bench2Drive-Plan, we extend
the study to NAVSIM, a benchmark that uses real-world images.

We adapt the privileged planners to NAVSIM v1.1~\citep{dauner2024navsim} and conduct camera
augmentation experiments using the same approach described above. Details of NAVSIM
adaptation and camera fusion are provided in
Appendices~\ref{app:navsim-implementation} and~\ref{app:camera}.

\begin{table}[H]
    \centering
    \footnotesize
    \setlength{\tabcolsep}{2pt}
    \renewcommand{\arraystretch}{0.96}
    \caption{Effect of front-view camera input. Base and +Camera denote configurations
    without and with the additional images; $\Delta$ is the score change. Positive changes
    are bold.}
    \label{tab:camera}
    \begin{tabular*}{\columnwidth}{@{\extracolsep{\fill}}lccc|ccc@{}}
        \toprule
        & \multicolumn{3}{c|}{Bench2Drive (DS $\uparrow$)}
        & \multicolumn{3}{c}{NAVSIM (PDMS $\uparrow$)} \\
        \cmidrule(lr){2-4} \cmidrule(lr){5-7}
        Method
        & Base & +Camera & $\Delta$DS
        & Base & +Camera & $\Delta$PDMS \\
        \midrule
        PLUTO             & 64.74 & 62.60 & -2.14 & 87.88 & 89.49 & \textbf{+1.61} \\
        Diffusion Planner & 58.78 & 58.05 & -0.73 & 87.51 & 88.36 & \textbf{+0.85} \\
        BEV-RGB           & 51.10 & 48.30 & -2.80 & 76.65 & 75.46 & -1.19 \\
        BEV-Mask          & 41.30 & 41.76 & \textbf{+0.46}
                          & 64.14 & 71.02 & \textbf{+6.88} \\
        InstructDriver    & 57.16 & 57.10 & -0.06 & 77.66 & 77.19 & -0.47 \\
        \bottomrule
    \end{tabular*}
\end{table}

Table~\ref{tab:camera} summarizes the results on both benchmarks. Camera inputs provide no
consistent gains on Bench2Drive-Plan but improve three of the five planners on NAVSIM. These
results suggest that additional visual information may benefit planning, although the gains vary across planners. Also, We
leave it to future work to investigate whether stronger fusion mechanisms or explicit
visual supervision would change the magnitude of these gains.

\section{Conclusion}

In this work, we introduce \textbf{Bench2Drive-Plan}, a unified benchmark that
bridges privileged planning and end-to-end autonomous driving under a shared
protocol. Through the shared scene preprocessing pipeline and planner adapters, Bench2Drive-Plan
enables privileged planners with diverse architectures and input representations to be
evaluated within the same closed-loop driving environment.
Our cross-paradigm experiments On Bench2Drive-Plan show that ground-truth scene inputs do not confer a large inherent
performance advantage over learned perception. We further study the role of additional
camera
images and observe no overall improvement on Bench2Drive-Plan, but an overall gain on NAVSIM. We hope
Bench2Drive-Plan can help the privileged planning and end-to-end driving communities share methods and learn from each other.



\FloatBarrier
\label{sec:main-content-end}
{\small
\bibliographystyle{ieeenat_fullname}
\bibliography{main}
}

\clearpage
\appendix
\maketitlesupplementary
\raggedbottom

\section{Training Hyperparameters}
\label{app:implementation}

Tables~\ref{tab:b2d}, \ref{tab:camera-train}, \ref{tab:hyper}, and
\ref{tab:navsim-camera-train}
summarize the training hyperparameters of each planner on Bench2Drive and NAVSIM, the
camera-augmented
Bench2Drive setting and the camera-augmented NAVSIM setting, respectively. Batch denotes
the
effective batch size. All models are trained in FP32 with gradient clipping at $5.0$. On
NAVSIM, BEV-RGB and BEV-Mask train the head and backbone with separate learning rates.

\begin{table*}[!t]
\centering
\caption{Training hyperparameters on Bench2Drive. LR denotes the initial learning rate; WD
denotes
weight decay.}
\label{tab:b2d}
{\small
\begin{tabular}{lcccccc}
\toprule
Method & Epochs & Batch & Optimizer & LR & WD & Schedule \\
\midrule
PLUTO & 25 & 128 & AdamW & $10^{-3}$ & $10^{-4}$ & 3-epoch warmup + cosine \\
Diffusion Planner & 400 & 2048 & AdamW & $5\times10^{-4}$ & $10^{-2}$ & 5-epoch warmup +
cosine \\
BEV-RGB & 25 & 128 & Adam & $10^{-3}$ & $10^{-4}$ & constant \\
BEV-Mask & 25 & 128 & Adam & $10^{-3}$ & $10^{-4}$ & constant \\
LLM planner & 3 & 8 & AdamW & $10^{-4}$ & $0$ & cosine + 0.1 warmup \\
\bottomrule
\end{tabular}%
}
\end{table*}

\begin{table*}[!t]
\centering
\caption{Training hyperparameters for the camera-augmented Bench2Drive setting. PLUTO,
Diffusion
Planner, and the LLM planner follow a two-stage schedule. Stage~1 freezes the ResNet-34 
backbone. Stage~2 unfreezes the backbone and jointly
fine-tunes the camera branch and planner. BEV-RGB and BEV-Mask use a single-stage schedule
that
jointly trains their unfrozen ResNet-50 BEV and ResNet-34 camera encoders. All camera
backbones are
initialized from ImageNet-pretrained weights.}
\label{tab:camera-train}
{\small
\begin{tabular}{lccccc}
\toprule
Method & Stage & Epochs & Batch & LR & Backbone \\
\midrule
PLUTO & 1 & 25 & 128 & $10^{-3}$ & ResNet-34 (frozen) \\
 & 2 & 5 & 128 & $10^{-4}$ & ResNet-34 (unfrozen) \\
\midrule
Diffusion Planner & 1 & 400 & 2048 & $5\times10^{-4}$ & ResNet-34 (frozen) \\
 & 2 & 100 & 128 & $10^{-4}$ & ResNet-34 (unfrozen) \\
\midrule
BEV-RGB & single & 25 & 64 & $10^{-3}$ & ResNet-34 (unfrozen) \\
\midrule
BEV-Mask & single & 25 & 64 & $10^{-3}$ & ResNet-34 (unfrozen) \\
\midrule
LLM planner & 1 & 3 & 8 & $10^{-4}$ & ResNet-34 (frozen) \\
 & 2 & 1 & 16 & $10^{-4}$ & ResNet-34 (unfrozen) \\
\bottomrule
\end{tabular}%
}
\end{table*}

\begin{table*}[!t]
\centering
\caption{Training hyperparameters for the trainable NAVSIM planners. LR denotes the
initial
learning rate; WD denotes weight decay.}
\label{tab:hyper}
{\small
\begin{tabular}{lcccccc}
\toprule
Method & Epochs & Batch & Optimizer & LR & WD & Schedule \\
\midrule
PLUTO & 50 & 32 & AdamW & $10^{-3}$ & $10^{-4}$ & 3-epoch warmup + cosine \\
Diffusion Planner & 72 & 64 & AdamW & $10^{-4}$ & $10^{-2}$ & 5-epoch warmup + cosine \\
BEV-RGB & 50 & 128 & AdamW & $10^{-4}/10^{-5}$ & $10^{-4}$ & 3-epoch warmup + cosine \\
BEV-Mask & 50 & 64 & AdamW & $10^{-4}/10^{-5}$ & $10^{-4}$ & 3-epoch warmup + cosine \\
LLM planner & 3 & 8 & AdamW & $10^{-4}$ & $0$ & cosine + 0.1 warmup \\
\bottomrule
\end{tabular}%
}
\end{table*}

\begin{table*}[!t]
\centering
\caption{Training hyperparameters for the camera-augmented NAVSIM setting. PLUTO,
Diffusion
Planner, and the LLM planner follow a two-stage schedule. Stage~1 freezes the
 ResNet-34 backbone. Stage~2 unfreezes the backbone and jointly
fine-tunes the camera branch and planner. BEV-RGB and BEV-Mask use a single-stage schedule
that
jointly trains their unfrozen ResNet-50 BEV and ResNet-34 camera encoders.
For BEV-RGB and BEV-Mask, the trajectory head is trained at the higher learning rate and
the two
backbones at the lower rate. All camera backbones are initialized from ImageNet-pretrained
weights.}
\label{tab:navsim-camera-train}
{\small
\begin{tabular}{lccccc}
\toprule
Method & Stage & Epochs & Batch & LR & Backbone \\
\midrule
PLUTO & 1 & 50 & 32 & $10^{-3}$ & ResNet-34 (frozen) \\
 & 2 & 10 & 32 & $10^{-4}$ & ResNet-34 (unfrozen) \\
\midrule
Diffusion Planner & 1 & 60 & 64 & $10^{-4}$ & ResNet-34 (frozen) \\
 & 2 & 15 & 16 & $2\times10^{-5}$ & ResNet-34 (unfrozen) \\
\midrule
BEV-RGB & single & 50 & 128 & $10^{-4}/10^{-5}$ & ResNet-34 (unfrozen) \\
\midrule
BEV-Mask & single & 50 & 64 & $10^{-4}/10^{-5}$ & ResNet-34 (unfrozen) \\
\midrule
LLM planner & 1 & 3 & 8 & $10^{-4}$ & ResNet-34 (frozen) \\
 & 2 & 1 & 16 & $10^{-4}$ & ResNet-34 (unfrozen) \\
\bottomrule
\end{tabular}%
}
\end{table*}

\section{Bench2Drive Implementation Details}
\label{app:bench2drive-implementation}

\subsection{Scene Feature Construction and planner adaptation}

Bench2Drive driving information are preprocessed into per-frame feature files for
training. During
closed-loop evaluation, the agent reconstructs the same feature definitions online from
the current
CARLA state. Both paths use a right-handed coordinate convention and an SE(2) transform
anchored
at the current ego pose. Actor identities are associated across frames, and future frames
are accessed
only when constructing training targets.

\paragraph{Traffic participants.}
We first extract ground-truth position, velocity and category for the ego vehicle and
traffic participants from the Bench2Drive annotations, and associate observations across
frames
using persistent actor identifiers. All states are transformed into a right-handed SE(2)
frame
anchored at the current ego pose. Benchmark-specific categories, including ordinary
vehicles,
pedestrians, two-wheelers, emergency vehicles, and vehicles with an open door, are
retained in the
shared scene representation. Missing history is padded and marked by validity masks, while
static
objects are recorded separately from dynamic actors. Each planner adapter then converts
these
shared states into its model-specific input format, using the required number of actors,
history
sampling scheme, and feature fields.

The Diffusion Planner retains the 32 nearest non-ego actors. Each actor is represented
over 21
frames (two seconds of history and the current frame at 10\,Hz) by ego-relative position,
heading
sine and cosine, velocity, footprint, and a five-way category encoding. Its training
targets
contain 80 future ego poses and trajectories for up to 10 neighboring agents over eight
seconds,
together with explicit validity masks. Five nearby static objects are represented by pose,
footprint, and type.

The PLUTO input retains up to 64 actors on the same 21-history/80-future timeline. It
stores
position, heading, velocity, footprint, category, and validity fields in PLUTO's
agent-feature
schema. Its current ego state additionally includes speed, acceleration, steering, and yaw
rate.
Future actor states are retained for training targets and reference-line projection only.

The two raster baselines render the scene into an ego-centered $300\times300$ raster at
5\,px/m
($0.2$\,m/px, a $60$\,m field of view). Actor and traffic rule geometry is drawn directly
from
benchmark annotations. BEV-RGB distinguishes semantic categories by color and encodes the
full
21-frame history through brightness, whereas BEV-Mask samples five frames and preserves
each
category and time step in a separate binary channel. Both models regress the 80-pose
trajectory
over eight seconds.

The LLM planner serializes up to 10 nearest dynamic agents as text, each described by a
category
(vehicle, walker, two-wheeler, emergency vehicle, or open-door vehicle), bounding-box
shape,
ego-relative pose, velocity, and a one-second position prediction, with static obstacles
listed
separately. The ego is described by its shape, pose, velocity, and acceleration together
with a
four-pose (two-second) historical trajectory; For prediction, the model decodes the
ego-relative 16-pose trajectory
over eight seconds as text.

\paragraph{Bench2Drive map preprocessing.}
For each town, we preprocess the Bench2Drive map into separate collections of lane
and
crosswalk objects. For each lane, fragmented centerline and boundary segments are merged
according
to endpoint proximity. Boundary segments are classified as left or right relative to the
centerline, aligned to a consistent direction, and combined into a lane polygon while
retaining
the lane topology. Crosswalk records are converted into polygons. Separate STRtree indices
are
built for the lane and crosswalk collections, enabling efficient spatial queries within
the local
map region. Dense route points are matched to their containing lane polygons, and the lane
topology constrains transitions to recover an ordered route-lane sequence.

Speed limits and traffic lights are associated with lanes in separate steps. Town-level
speed-limit samples are converted to the right-handed map frame and indexed with a
cKDTree. For
each queried lane, we retrieve speed-limit samples within 3\,m of its centerline and use
their
median value. At each frame, we select traffic-light annotations marked as affecting the
ego
vehicle, query each trigger location against the lane STRtree, and write its one-hot light
state
into the corresponding lane tokens.

\paragraph{Ego-centered local-window extraction.}
\label{app:lane_cropping}
Most privileged planners evaluated in this work were originally developed on nuPlan, but
the lane
length distribution differs substantially between nuPlan and Bench2Drive. In our processed maps,
$22.3\%$ of Bench2Drive lanes exceed 100\,m, compared with $5.0\%$ in nuPlan; above
200\,m, the
proportions are $12.1\%$ and $0.8\%$, respectively. This gap is important for planners
that encode
each lane with a fixed budget of 20 points. Uniformly sampling a fixed number of points
over an
entire long Bench2Drive lane places many samples far from the ego vehicle and represents the
local road
geometry too sparsely. This differs from the denser local lane representation typically
encountered
in nuPlan.

\begin{figure*}[!t]
  \centering
  \includegraphics[width=0.92\textwidth]{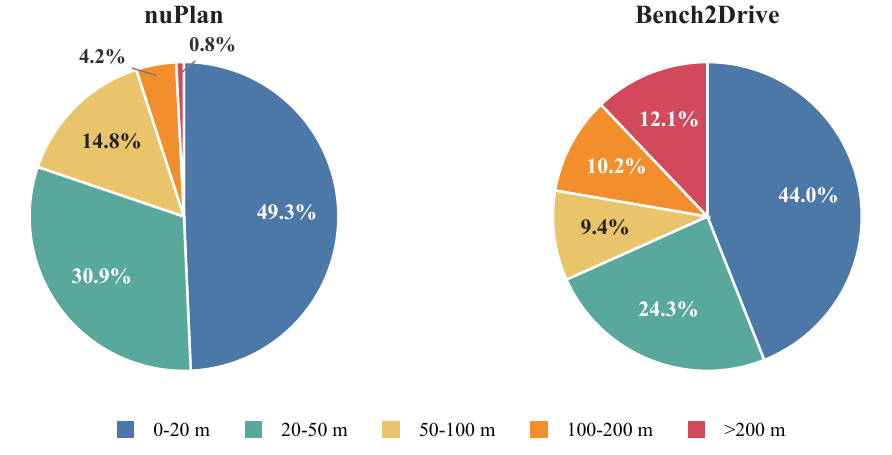}
  \caption{\textbf{Distribution of lane lengths in nuPlan and Bench2Drive.} The
  substantial
  difference in the prevalence of long lanes between Bench2Drive and nuPlan motivates our
  ego-centered
  local-window extraction procedure.}
  \label{fig:lane-length-distribution}
\end{figure*}

We address this gap with an ego-centered local-window procedure. For each queried lane, we
intersect its centerline with the local map region and select the connected segment
closest to the
ego vehicle. If the intersection does not yield a valid segment, we project the ego
position onto
the complete centerline and construct a local interval around the projected arc length. We
convert
the start and end positions of the selected segment into normalized progress along the
complete
centerline. The same progress interval is then used to crop and resample the centerline
and both
lane boundaries into 20 aligned points.
This preserves the correspondence between the centerline and boundaries while
concentrating the
fixed point budget on geometry relevant to the current planning context. In our Diffusion
Planner
experiment, replacing whole-lane sampling with this procedure substantially improved
performance
on Bench2Drive, showing the importance of correcting this map representation gap when
transferring
planners from nuPlan to CARLA.

After preprocessing, each adapter combines the extracted map information with the
remaining scene features and converts them into the input format required by its planner.

The Diffusion adapter orders local map elements by distance and retains up to 70 lane or
crosswalk
tokens and 25 route-lane tokens, with 20 points per token. Lane tokens contain centerline
position
and direction, boundary offsets, traffic-light state, lane/crosswalk type, and speed-limit
value and
validity; route-lane tokens reuse the corresponding ordered local-lane geometry.

The PLUTO adapter resamples each lane into 20 aligned segments along its centerline and
two
boundaries, while crosswalks use the same point sequence format. Each map element contains
point
positions, direction vectors, orientations, and side labels, together with polygon
attributes for
map type, route membership, traffic light state, and road block identity. Candidate
reference
lines are constructed by starting from nearby route lanes and traversing the route
topology. They
are limited to 100\,m, duplicate candidates are removed, and the remaining lines are
sampled at
1\,m intervals. Future ego poses are projected onto each candidate to provide progress and
distance
targets for reference line supervision. To form the cost map, the adapter rasterizes
drivable lane
polygons, removes stationary actor footprints, and converts the result into a signed
distance field.

For the raster baselines, we first rasterize each Bench2Drive town map into a road surface layer
and a
lane geometry layer containing lane centerlines and boundaries. At each frame, both layers
are
cropped and rotated around the current ego pose to form a local BEV representation. We
then add the
near and far route command points, stop signs, and the stop lines of traffic lights
affecting the
ego vehicle from the current annotations.

BEV-RGB stores these map elements in one color image: road, route, and lane geometry are
rendered
in dark gray, light gray, and magenta, while traffic light stop lines use green, yellow,
or red and
stop signs use dark yellow. BEV-Mask instead represents road, route, and lane geometry
with three
binary channels. Green traffic lights, combined red and yellow traffic lights, and stop
signs each
use five temporal channels.

The LLM planner serializes the ego's current lane centerline and its left and right
boundaries,
each sampled forward up to 100\,m at ten points, together with the current traffic-light
status,
the speed limit, and the upcoming stop line as a line segment with its distance.

\subsection{Planner Implementations and Training Configurations}

\paragraph{Diffusion Planner.}
We adapt the Diffusion Planner implementation \citep{zheng2025diffusion} to
Bench2Drive while retaining its overall architecture and planning formulation. Bench2Drive
contains
more diverse interactive scenarios and semantic categories than the original planner
input. For
example, the \textit{YieldToEmergencyVehicle} scenario requires the planner to distinguish
an
emergency vehicle from an ordinary vehicle. We therefore adapt the encoder to preserve
these
task-relevant semantics. The neighbor type embedding is expanded from three to five
categories:
vehicles, pedestrians, two-wheelers, emergency vehicles, and open-door vehicles. We also
add a
binary lane/crosswalk type embedding to the lane encoder, while leaving the remaining
model
components unchanged. 

We train the adapted model for 400 epochs with
four-GPU distributed data parallelism, a per-GPU batch size of 512, AdamW at an initial
learning rate of
$5\times10^{-4}$, five warmup epochs followed by cosine decay. Encoder and
decoder drop-path rates are both 0.1. An exponential moving average of model parameters is
maintained with decay 0.999.
We retain the original Diffusion Planner loss and state perturbation augmentation.

\paragraph{PLUTO.}
We adapt the PLUTO implementation \citep{cheng2024pluto} to Bench2Drive while retaining
its
agent and map encoders, reference-line-conditioned planning decoder, trajectory-mode head,
and
associated planning losses. PLUTO faces the same actor-category mismatch as Diffusion
Planner.
We therefore expand its agent type embedding from four to eight entries. All other model
components
follow the original implementation. 

We train the adapted model for 25 epochs on a single GPU with a batch size of 128,
using AdamW at an initial learning rate of
$10^{-3}$ and weight decay $10^{-4}$, with a three-epoch linear warmup followed by cosine
decay to
$10^{-6}$. It predicts 12 trajectory modes per reference line from a 21-frame (two-second)
history
and outputs 80 poses over eight seconds.

\paragraph{BEV-RGB and BEV-Mask.}
Both raster baselines use a $300\times300$ ego-centered input at $0.2$\,m/px,
corresponding to a
$60$\,m field of view. BEV-RGB combines map, route, traffic rules, actors, and the ego
vehicle in a
three-channel semantic image. Actor categories are distinguished by color, and the
complete
21-frame history is encoded by shifting older footprints toward white while retaining the
full
category color at the current frame. The image is scaled to $[0,1]$ and normalized with
ImageNet
statistics. BEV-Mask represents the scene as a 38-channel binary tensor. Three channels
encode
road, route, and lane geometry. Vehicles, including the ego vehicle and open-door
indicators,
emergency vehicles, static objects, and walkers each use five channels sampled uniformly
from the
21-frame history. Green traffic lights, combined red and
yellow traffic lights, and stop signs use another five channels each. Missing history is
padded
with zeros.

Both models use an ImageNet-pretrained ResNet-50 backbone. BEV-RGB retains the standard
three-channel input stem, whereas the input convolution of BEV-Mask is expanded to 38
channels.
The remaining backbone structure is identical. After global average pooling, a linear
trajectory
head outputs 240 values, which are reshaped into 80 future ego poses over eight seconds,
with each
pose represented by $(x,y,\mathrm{heading})$. We train both models for 25 epochs with
Adam, a batch
size of 128, an initial learning rate of $10^{-3}$, weight decay of $10^{-4}$. The
imitation loss combines
position MSE and wrapped angular L1 loss for heading with
equal weights.

\paragraph{InstructDriver.}
We adapt InstructDriver \citep{zhang2024instruct} to Bench2Drive while preserving its
Perception, Ego-States, Historical Trajectory, and Map Insights prompt structure and LoRA
recipe~\citep{hu2021lora}.
We serialize the Bench2Drive scene representation into the corresponding prompt fields and
use 16 ego-relative poses as trajectory targets. Qwen2.5-1.5B-Instruct is fine-tuned for
three epochs with rank-8 LoRA applied
to all linear modules~\citep{bai2023qwen}. At inference, greedy decoding (temperature $0$,
maximum generation length
$3072$) produces the $(x, y, \mathrm{yaw})$ trajectory as text, which is converted to the
Bench2Drive trajectory interface.

\subsection{Implementation of the PLUTO--VAD Study}
\label{app:cross-paradigm-details}

\noindent\textbf{Scene interfaces and architectural variants.}
PLUTO--VAD uses VAD's image backbone and temporal BEV network to learn object, map,
static-obstacle, and reference-line queries. Object/map/static features are projected from
256 to 128 dimensions, concatenated with an ego-state token, and processed by four
scene-interaction layers. Reference-query features condition a PLUTO-style multimodal
planning decoder. Thus, camera-derived queries supply the scene representation previously
obtained from PLUTO's structured scene encoders.

No-reference removes the reference prediction branch and its supervision, the reference
encodings, and reference-to-reference attention. A learned embedding of the near
navigation command is concatenated with each of six mode embeddings and projected to
initialize the trajectory queries. Each planning layer retains mode self-attention, scene
cross-attention, and a feed-forward network. Separate heads predict XY positions, heading
as sine/cosine, velocity, and mode scores. Training selects the mode with the lowest
masked trajectory regression cost and supervises its score with cross-entropy. The model
retains 300 object queries, 100 map queries, 20 static-obstacle queries, and PLUTO's
AgentPredictor.

Motion replaces AgentPredictor with VAD's motion processing and uses its fused agent
features as planning context. It uses 320 unified object queries and 100 map queries, with
static objects included in detection, while retaining the navigation-conditioned
multimodal planning structure. Endpoint uses VAD motion features for agents and a lane
encoder for map features. A learned ego query first attends to agent features; the
resulting query then attends to map features. Both interaction outputs are concatenated
and passed to a trajectory MLP, producing six command-specific trajectories. Navigation
selects one branch. Endpoint does not retain the PLUTO-style four-layer scene encoder or
the six planning-mode queries. The sensor variants predict 30 future points over three
seconds. No-reference and Motion mask low-confidence scene tokens at a threshold of 0.5;
the baseline Endpoint uses threshold zero.

Endpoint follows the Bench2Drive category definitions used in this study, including
distinctions for two-wheelers and emergency vehicles. During closed-loop evaluation, its
navigation command is obtained from the near waypoint returned by the route planner and
mapped to the six command indices.

\noindent\textbf{Displacements and positions.}
Let $\mathbf{y}_{bt}$ denote the future ego position, expressed relative to the current
ego pose, with $\mathbf{y}_{b0}=\mathbf{0}$. The baseline Endpoint predicts displacements
$\hat{\boldsymbol\delta}_{bct}$ for command $c$, with targets
$\boldsymbol\delta_{bt}=\mathbf{y}_{bt}-\mathbf{y}_{b,t-1}$. At inference, positions are
recovered by cumulative summation. The position variant directly predicts
$\hat{\mathbf{y}}_{bct}$ and supervises it against $\mathbf{y}_{bt}$; inference applies no
cumulative summation. Positions are local ego-frame coordinates, not global coordinates.
The native LiDAR ordering $(\mathrm{right},\mathrm{forward})$ is converted once to the
planner/controller ordering $(\mathrm{forward},\mathrm{left})$.

\noindent\textbf{Effective planning-loss scale.}
For the position-predicting variants, let $c_b$ be the navigation command and $v_{bt}$ the
future validity mask. Define the masked regression sum
\begin{equation}
S=\sum_{b=1}^{B}\sum_{t=1}^{T}v_{bt}
\|\hat{\mathbf{y}}_{b,c_b,t}-\mathbf{y}_{bt}\|_1.
\end{equation}
With $K=6$ command branches and $T=30$ steps, the original weighted mean is
$\mathcal{L}_{\mathrm{original}}=S/(2BKT)$. Unselected branches and invalid steps
contribute zero loss but remain in the denominator. We adjust the effective scale by using
\begin{equation}
\mathcal{L}_{\mathrm{adjusted}}=
\frac{S}{\max(2\sum_{b,t}v_{bt},1)}.
\end{equation}
The nominal planning-regression coefficient remains one; the adjustment is entirely
through normalization. With all time steps valid, it increases this loss by a factor of
six, and the factor is larger when some steps are invalid. Under distributed training, the
valid-coordinate count is aggregated over all workers and the local loss is scaled to
account for gradient averaging, yielding the global valid-coordinate mean. Loss magnitude
alone does not determine the task's gradient contribution; this experiment evaluates the
effect of the normalization change through closed-loop performance.

\noindent\textbf{Confidence filtering.}
For query $q$ with final classification logits $\mathbf{z}_q$, its confidence is
$s_q=\max_j\sigma(z_{qj})$. We retain queries with $s_q\geq0.5$ for the corresponding ego
attention. Agent features are taken after motion processing, while their scores come from
the associated detection classification head. Map scores come from the map classification
head. Variable query counts are padded across the batch and padding is masked. If a sample
has no retained query, a zero-feature, zero-position token is left unmasked to avoid an
empty attention context. This selection affects ego planning interactions; perception and
motion supervision retain their existing query assignments.

\subsection{Closed-Loop Execution}

All methods are evaluated on the official 220-route Bench2Drive split through CARLA
Leaderboard
agents. The predicted ego trajectory is transformed once from the model's right-handed ego
frame
to CARLA's left-handed local frame. All such methods then use the official Bench2Drive
trajectory-tracking controller with
identical parameters for every planner.

Given predicted waypoints $\{p_i\}_{i=1}^{N}$, the controller computes the desired speed
as
$v_{\mathrm{des}}=\frac{10}{N-1}\sum_i\|p_{i+1}-p_i\|_2$ and selects the waypoint
associated with
the segment midpoint nearest 4\,m as the steering aim point. The lateral PID gains
$(K_P,K_I,K_D)$ are $(0.75,0.75,0.3)$, while the longitudinal gains are $(5.0,0.5,1.0)$;
both use
a 40-step error window. Braking is activated when the desired speed is below
$0.4\,\mathrm{m/s}$
or the current-to-desired speed ratio exceeds 1.1. The longitudinal speed error is clipped
at 0.25
and throttle at 0.75. Final steering, throttle, and brake commands are clipped to
$[-1,1]$,
$[0,0.75]$, and $[0,1]$, respectively, and throttle is suppressed above an ego speed of
$5\,\mathrm{m/s}$.

\section{NAVSIM Implementation Details}
\label{app:navsim-implementation}

\subsection{Scene Feature Construction and Planner Adaptation}

NAVSIM scenes are converted into planner inputs from the dataset annotations and map API.
All
geometric features are expressed in a local SE(2) frame anchored at the current ego pose.
The
standard observation contains four recorded history frames. Future frames are used only to
build
training targets. Each planner outputs eight ego-relative poses at $0.5$-second intervals
over four
seconds, following the NAVSIM trajectory interface.

\paragraph{Traffic participants and map information.}
We extract the ego state, surrounding actors, local map geometry, route, traffic-light
states, and
speed limits from each scene. Each adapter then organizes this information according to
the actor
capacity, temporal sampling, and feature definitions required by its planner.

The Diffusion Planner input contains the current ego state, four-frame histories for up to
32
neighboring agents, and five static objects. Its vectorized map input retains up to 70
local lanes
and 25 route lanes, with 20 sampled points per polyline, together with traffic-light and
speed-limit
attributes. The training targets contain the future ego trajectory and trajectories for up
to 10
neighboring agents.

The PLUTO input retains up to 48 actors and 10 static obstacles. Lane and route geometry
is
converted into polygon features with 20 sampled points, and the adapter additionally
constructs
candidate reference lines and the cost map used by PLUTO's planning objective.

The two raster baselines render the four-frame scene history into an ego-centered
$300\times300$ BEV at 5\,px/m. BEV-RGB represents the map, route, traffic rules, actors,
and ego
vehicle in a three-channel semantic image. Actor classes are distinguished by color, and
arrows
encode current velocities. BEV-Mask represents the same scene information in 24 binary
channels.
Three channels encode road, route, and lane geometry; four temporal channels are used for
each of
vehicles, static objects, walkers, green lights, and red lights; and one channel
represents stop
signs.

The LLM planner follows the InstructDriver prompt structure \citep{zhang2024instruct}. The
adapter
serializes the ego state, up to 10 surrounding actors, and nearby map and traffic
information
into an ego-centric text prompt. The model generates eight ego-relative
$(x,y,\mathrm{yaw})$ poses
as text, which are parsed into the NAVSIM trajectory format.

\subsection{Planner Implementations and Training Configurations}

\paragraph{Diffusion Planner.}
We adapt the Diffusion Planner implementation \citep{zheng2025diffusion} to NAVSIM while
retaining
its overall architecture and planning formulation. The adapter reuses the original agent,
static
object, lane, and route encoders, scene fusion, diffusion decoder, and sampler. We retain
the
original state perturbation augmentation and adapt the training loss to the NAVSIM
setting.

We train the adapted model for 72 epochs with AdamW, a batch size of 64, and an initial
learning rate of $10^{-4}$. The learning rate uses five warmup epochs followed by cosine
decay to
$10^{-5}$. An exponential moving average of the model parameters is maintained with decay
$0.999$.

\paragraph{PLUTO.}
We adapt PLUTO \citep{cheng2024pluto} while retaining its map encoder, planning decoder,
trajectory-mode head, and planning losses. Four NAVSIM history frames yield only three
temporal
difference tokens, which are too short for the neighborhood kernel in PLUTO's original
temporal
encoder. We therefore replace this block with a two-layer self-attention encoder and
retain the
remaining model components.

We train the adapted model for 50 epochs with AdamW, a batch size of 32, and an initial
learning rate of $10^{-3}$. The learning rate uses three warmup epochs followed by cosine
decay to
$10^{-6}$. We retain PLUTO's planning, neighboring-agent prediction, and collision losses,
together
with its positive-sample perturbation adapted to the NAVSIM input.

\paragraph{BEV-RGB and BEV-Mask.}
Both raster planners use an ImageNet-pretrained ResNet-50 backbone and a linear trajectory
head.
BEV-RGB retains the standard three-channel input stem, whereas the BEV-Mask input
convolution is
expanded to 24 channels. The head outputs 24 values, which are reshaped into eight future
$(x,y,\mathrm{heading})$ poses.

We train both models for 50 epochs with AdamW and three warmup epochs followed by cosine
decay.
BEV-RGB and BEV-Mask use batch sizes of 128 and 64, respectively. For both models, the
trajectory head uses an initial learning rate of $10^{-4}$ and the pretrained backbone
uses
$10^{-5}$. Their shared imitation loss combines position MSE and wrapped angular L1 loss
for
heading with equal weights.

\paragraph{InstructDriver.}
We fine-tune Qwen2.5-1.5B-Instruct with rank-8 LoRA applied to all linear modules.
Training uses the
token-level supervised fine-tuning objective and the Qwen chat template for three epochs,
with an
batch size of 8, an initial learning rate of $10^{-4}$, and a cosine schedule with a
warmup ratio of $0.1$. At evaluation time, greedy decoding generates at most 1024 new
tokens before
the predicted poses are parsed from the response.

\subsection{Trajectory Evaluation}

All methods are evaluated on the 12,146 scenarios of the official navtest split. For each
scene,
the adapter reconstructs the same input representation used during training and returns
eight
ego-relative poses spanning four seconds.

The NAVSIM v1.1 scorer interpolates the 2\,Hz planner output to a 10\,Hz, 40-pose proposal
and runs
PDM simulation under the fixed logged scene evolution.
We use a shared precomputed metric cache and report the official PDM Score (PDMS),
together with
no-at-fault collision, drivable-area compliance, ego progress, time-to-collision, and
comfort.

\section{Camera Augmentation Details}
\label{app:camera}

\subsection{Visual Encoder and Fusion}

All planners
follow the encoder-level concatenation protocol of Sec.~\ref{sec:camera-study}. The
front-view RGB
image is resized to $900\times256$ pixels and encoded by an ImageNet-pretrained ResNet-34,
and a
learned projection (a linear layer followed by GELU and LayerNorm) maps the ResNet output
into the
planner's feature space.
Image normalization uses the ImageNet mean and standard deviation in both training and
evaluation.
The projected camera feature is then combined with each planner's scene representation.
Section~\ref{app:camera-specific} describes the model-specific fusion designs, and
Table~\ref{tab:camera-train} reports the training hyperparameters.

\subsection{Planner-Specific Camera Fusion}
\label{app:camera-specific}

\paragraph{Diffusion Planner.}
Diffusion Planner first encodes the front-view image with ResNet-34. A linear layer
followed by
GELU and LayerNorm maps the 1000-dimensional image feature to the planner's hidden
dimension. The
resulting camera token is scaled by a learnable factor initialized to 0.1 and appended to
the
encoded agent, static-object, and lane tokens before the scene fusion Transformer.
Training uses two stages. Stage~1 trains the projection and planner on precomputed
features from a
frozen ResNet-34 for 400 epochs, with batch size 512 and learning rate $5\times10^{-4}$.
Stage~2
loads the first-stage weights, unfreezes the ResNet-34, and jointly fine-tunes the
complete model on
raw images for 100 epochs, with batch size 128 and learning rate $10^{-4}$. Both stages
use four
GPUs, five warmup epochs followed by cosine decay, and EMA decay 0.999.

\paragraph{PLUTO.}
PLUTO also encodes the front-view image with ResNet-34. A linear layer followed by GELU
and
LayerNorm maps the image feature to PLUTO's hidden dimension. The resulting camera token
is scaled
by a learnable factor initialized to 0.1 and appended to the agent, map, and
reference-line tokens
before the fusion Transformer. Training uses two stages. Stage~1 trains the projection and
planner
on precomputed features from a frozen ResNet-34 for 25 epochs, with batch size 128 and
learning rate
$10^{-3}$. Stage~2 unfreezes the ResNet-34 and jointly fine-tunes the complete model on
raw images
for five epochs, with batch size 128 and learning rate $10^{-4}$.

\paragraph{BEV-RGB and BEV-Mask.}
BEV-RGB and BEV-Mask encode the front-view image into a 512-dimensional feature using
ResNet-34. A
linear layer followed by GELU and LayerNorm maps this output to a 256-dimensional camera
feature.
This feature is concatenated with the global feature produced by the ResNet-50 BEV
encoder. A fusion MLP combines the two features before the trajectory head. The ResNet-34,
ResNet-50, fusion module, and trajectory
head are jointly trained on raw images for 25 epochs, with batch size 64 and learning rate
$10^{-3}$.

\paragraph{InstructDriver.}
InstructDriver encodes the front-view image with ResNet-34. A linear layer
followed by GELU and LayerNorm maps the image feature to the hidden dimension of
Qwen2.5-1.5B-Instruct. The resulting camera token is scaled by a learnable factor
initialized to
0.1 and prepended to the text token sequence. Training uses two stages. Stage~1 trains the
LoRA
adapter on precomputed features from a frozen ResNet-34 for three epochs, with batch size
8 and
learning rate $10^{-4}$. Stage~2 unfreezes the ResNet-34 and jointly fine-tunes it with
the LoRA
adapter on raw images for one epoch, with batch size 16 and the same learning rate.

\section{Additional Results}
\label{app:additional-results}

\subsection{NAVSIM Diagnostic Results}
\label{app:navsim-diagnostics}

Tables~\ref{tab:navsim-full-base} and~\ref{tab:navsim-full-camera} report all NAVSIM
metrics before
and after camera augmentation. Scores use a 0--100 scale over 12,146 test scenes, and the
metrics
follow Sec.~4.1.

\begin{table*}[htbp]
\caption{Complete NAVSIM results without front-view camera input. The end-to-end results
are
officially reported NAVSIM references. Bold marks the highest score within each method
group~\citep{dauner2024navsim}.}
\label{tab:navsim-full-base}
\centering
\setlength{\tabcolsep}{5.2pt}
\renewcommand{\arraystretch}{1.10}
\begin{tabular}{@{}lcccccc@{}}
\toprule
Method & PDMS $\uparrow$ & NAC $\uparrow$ & DAC $\uparrow$ & EP $\uparrow$
& TTC $\uparrow$ & Comfort $\uparrow$ \\
\midrule
\multicolumn{7}{l}{\emph{Privileged planners}} \\
PLUTO & \textbf{87.88} & 97.43 & 97.33 & \textbf{82.18} & 93.39 & 99.93 \\
Diffusion Planner & 87.51 & \textbf{97.95} & \textbf{97.55} & 80.90
& \textbf{93.75} & 98.66 \\
BEV-RGB & 76.65 & 94.53 & 88.86 & 72.66 & 87.48 & 99.12 \\
BEV-Mask & 64.14 & 88.44 & 82.04 & 61.61 & 77.89 & 99.88 \\
InstructDriver & 77.66 & 96.63 & 88.30 & 71.03 & 91.03 & \textbf{99.99} \\
\midrule
\multicolumn{7}{l}{\emph{Official NAVSIM end-to-end references}} \\
LTF~\citep{chitta2022transfuser} & 83.80 & 97.40 & \textbf{92.80} & 79.00
& 92.40 & \textbf{100.00} \\
TransFuser~\citep{chitta2022transfuser} & \textbf{84.00} & 97.70 & \textbf{92.80}
& 79.20 & 92.80 & \textbf{100.00} \\
UniAD~\citep{hu2023planning} & 83.40 & 97.80 & 91.90 & 78.80 & 92.90
& \textbf{100.00} \\
PARA-Drive~\citep{weng2024drive} & \textbf{84.00} & \textbf{97.90} & 92.40
& \textbf{79.30} & \textbf{93.00} & 99.80 \\
\bottomrule
\end{tabular}
\end{table*}

PLUTO and Diffusion Planner achieve $87.88$ and $87.51$ PDMS, respectively, compared with
$84.00$ for the strongest included end-to-end reference. The difference is most visible
in drivable-area compliance and ego progress, while the best collision and TTC scores are
closer
between the two groups. Comfort is saturated for both groups. These results place the
leading
privileged and end-to-end methods in a similar performance range, while showing that
direct scene
states provide a moderate rather than decisive advantage on NAVSIM.

\begin{table*}[htbp]
\caption{Complete NAVSIM results with front-view camera input. Bold marks the highest
score in
each column.}
\label{tab:navsim-full-camera}
\centering
\setlength{\tabcolsep}{5.2pt}
\renewcommand{\arraystretch}{1.10}
\begin{tabular}{@{}lcccccc@{}}
\toprule
Method & PDMS $\uparrow$ & NAC $\uparrow$ & DAC $\uparrow$ & EP $\uparrow$
& TTC $\uparrow$ & Comfort $\uparrow$ \\
\midrule
PLUTO & \textbf{89.49} & \textbf{97.86} & \textbf{98.16} & \textbf{83.95}
& \textbf{94.20} & 99.96 \\
Diffusion Planner & 88.36 & 97.75 & 97.56 & 82.90 & 93.77 & 98.89 \\
BEV-RGB & 75.46 & 94.83 & 87.19 & 71.58 & 87.85 & 98.69 \\
BEV-Mask & 71.02 & 92.98 & 84.55 & 67.63 & 84.99 & 98.28 \\
InstructDriver & 77.19 & 95.71 & 87.55 & 72.19 & 89.55 & \textbf{99.98} \\
\bottomrule
\end{tabular}
\end{table*}

\begin{table*}[!t]

\caption{Driving ability scores (\%) on the 220-route Bench2Drive evaluation. The first
five methods are privileged planners, while the remaining six are sensor-based end-to-end
baselines. Mean is the unweighted average of the five abilities. $^\ast$ denotes
expert-feature distillation.~\citep{jia2024bench2drive}.}

\label{tab:b2d-ability}

\centering
\small

\setlength{\tabcolsep}{4pt}
\renewcommand{\arraystretch}{1.08}

\begin{tabular}{@{}lcccccc@{}}

\toprule
Method & Overtaking$\uparrow$ & Merging$\uparrow$ & Emergency Brake$\uparrow$ & Give
Way$\uparrow$ & Traffic Signs$\uparrow$ & Mean$\uparrow$ \\
\midrule

\rowcolor{gray!10}
PLUTO 
& \textbf{44.44} & \textbf{33.75} & \textbf{28.33} & \textbf{40.00} & 34.21 &
\textbf{36.15} \\

Diffusion Planner 
& 31.11 & 27.50 & 21.67 & 20.00 & 30.53 & 26.16 \\

\rowcolor{gray!10}
BEV-RGB 
& 13.33 & 25.00 & 18.33 & 20.00 & 31.05 & 21.54 \\

BEV-Mask 
& 17.78 & 16.25 & 8.33 & 10.00 & 21.58 & 14.79 \\

\rowcolor{gray!10}
InstructDriver 
& 17.78 & 25.00 & 23.33 & 20.00 & \textbf{37.37} & 24.70 \\

\midrule

VAD 
& 13.04 & 17.19 & 29.82 & 16.67 & 19.77 & 19.30 \\

\rowcolor{gray!10}
UniAD-Base 
& 11.59 & 10.94 & 40.35 & \textbf{33.33} & 8.14 & 20.87 \\

TCP-traj$^\ast$
& 14.49 & 14.06 & 59.64 & 33.33 & 63.95 & 37.10 \\

\rowcolor{gray!10}
ThinkTwice$^\ast$
& 20.29 & 20.31 & 59.64 & 33.33 & 66.86 & 40.09 \\

DriveTransformer
& \textbf{30.43} & \textbf{39.06} & 49.12 & 0.00 & 61.63 & 36.05 \\

\rowcolor{gray!10}
DriveAdapter$^\ast$
& 26.09 & 21.88 & \textbf{64.90} & \textbf{33.33} & \textbf{69.18} & \textbf{43.08} \\

\bottomrule

\end{tabular}

\end{table*}

Without camera input, PLUTO and Diffusion Planner form the leading group on NAVSIM, with
only
$0.37$ PDMS between them. Their component profiles nevertheless differ: Diffusion Planner
obtains
the highest NAC, DAC, and TTC, whereas PLUTO achieves higher ego progress and the best
overall
PDMS. InstructDriver and BEV-RGB occupy the middle group, while BEV-Mask remains lower
across most
safety and progress metrics. Comfort is nearly saturated for all five planners and
therefore
provides limited separation among them.

The overall NAVSIM ranking, PLUTO, Diffusion Planner, InstructDriver, BEV-RGB, and
BEV-Mask,
matches the Bench2Drive DS ranking. Camera augmentation does not change this ordering on
either
benchmark, but its numerical effect differs. On NAVSIM, PLUTO and Diffusion Planner
improve by
$1.61$ and $0.85$ PDMS, and BEV-Mask improves by $6.88$ PDMS, whereas BEV-RGB and
InstructDriver
decrease slightly. Bench2Drive shows little or negative change for most planners. The
repeated
ordering indicates a stable coarse performance trend across the two evaluations, while the
different camera gains show that additional visual information interacts with the dataset
and the
planner representation rather than providing a uniform benefit.

\subsection{Bench2Drive Scenario-Level Results}
\label{app:b2d-scenario-results}

\begin{figure*}[p]
\centering
\makebox[\textwidth][c]{%
\includegraphics[width=1.0\textwidth,height=0.90\textheight,keepaspectratio]
{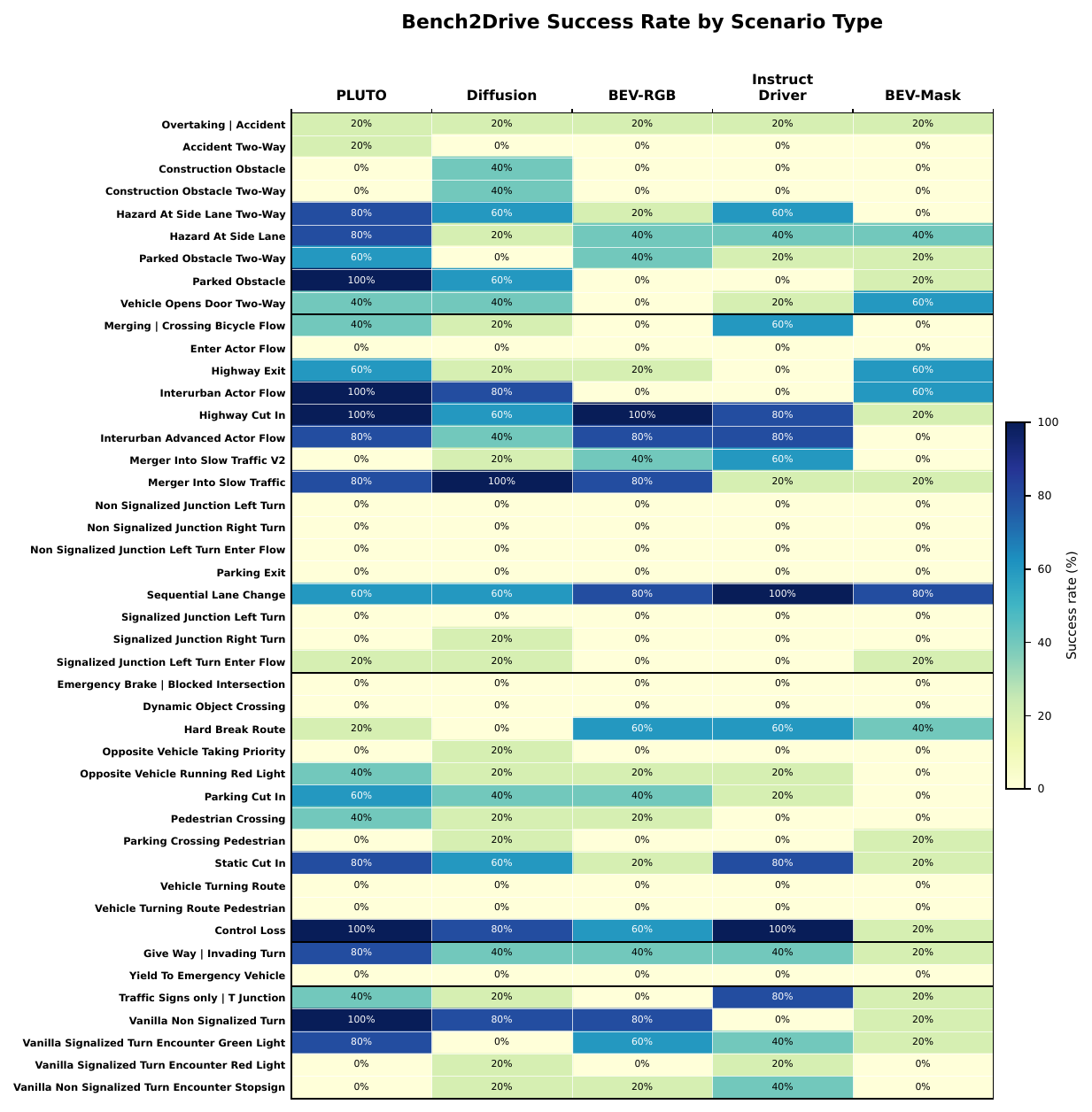}%
}
\caption{Scenario-level success rates on Bench2Drive. Each scenario contains five routes;
both
cell annotations and color encode the corresponding success rate.
Thick horizontal lines separate the primary display groups.}
\label{fig:b2d-scenario-sr}
\end{figure*}

\begin{figure*}[p]
\centering
\makebox[\textwidth][c]{%
\includegraphics[width=1.0\textwidth,height=0.90\textheight,keepaspectratio]
{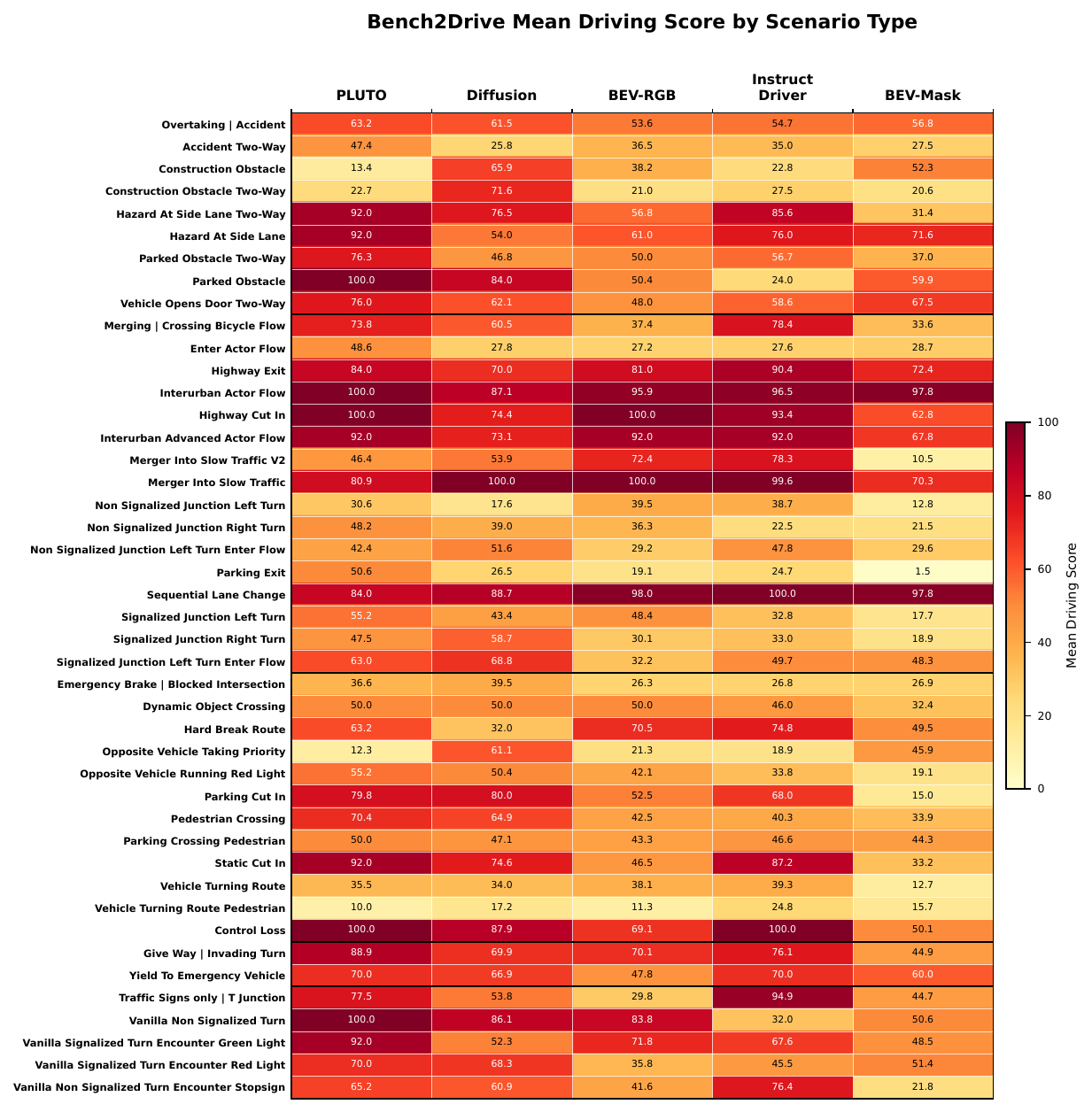}%
}
\caption{Mean Driving Score by Bench2Drive scenario type. Each cell averages DS over the
five
routes of the corresponding scenario and therefore retains both route completion and
infraction
penalties.}
\label{fig:b2d-scenario-ds}
\end{figure*}

The 220-route Bench2Drive evaluation contains five routes for each of 44 atomic scenario
types.
Following the official SR criterion, a route is successful only if its status is Completed
or
Perfect and it contains no infraction other than a minimum-speed infraction.
Figure~\ref{fig:b2d-scenario-sr} reports the scenario success rate in every cell.
Figure~\ref{fig:b2d-scenario-ds} reports the mean DS of the corresponding routes.

The heatmaps reveal a broad agreement in scenario difficulty. Across planner families,
highway and
interurban flow, highway cut-in, merger-into-slow-traffic, sequential-lane-change, and
vanilla-turn
scenarios tend to receive higher scores, whereas accident or construction setups,
non-signalized
junction interactions, parking exits, and emergency-yield cases remain challenging for
most methods.
Although individual cells differ across planners, the mean-DS and success-rate maps
exhibit broadly
similar trends in scenario difficulty.

\end{document}